\documentclass[11pt]{article}

\usepackage[utf8]{inputenc}
\usepackage[T1]{fontenc}
\usepackage{amsmath,amssymb,amsfonts}
\usepackage{graphicx}
\graphicspath{{./}{paper/}} 
\usepackage{booktabs}
\usepackage{tabularx}
\usepackage{multirow}
\usepackage{xcolor}
\usepackage{hyperref}
\usepackage[margin=1in]{geometry}
\usepackage{authblk}
\usepackage{enumitem}
\usepackage[numbers]{natbib}
\usepackage{tikz}
\usetikzlibrary{arrows.meta,positioning,fit,backgrounds,calc,decorations.pathreplacing}
\usepackage[font=small,labelfont=bf,skip=4pt]{caption}  

\hypersetup{colorlinks=true, linkcolor=blue, citecolor=blue, urlcolor=blue}

\newcommand{\runA}{\textsc{Dense}}     
\newcommand{\runP}{\textsc{Loop}}      
\newcommand{\runD}{\textsc{LMC}}       
\newcommand{\runM}{\textsc{Dense+Mem}} 
\newcommand{\runMiso}{\textsc{DenseMem-Iso}} 
\newcommand{\runPbig}{\textsc{Loop-Iso}}     
\newcommand{\runDiso}{\textsc{LMC-Iso}}      
\newcommand{\runDsmall}{\textsc{LMC-13}}     
\newcommand{\runDowt}{\textsc{LMC-OWT}}      

\title{\bfseries Repeated Shared Access Enables Grokking,\\
       but Edit Propagation Depends on an\\
       Addressable Memory}
\author{Yanan Niu\thanks{Correspondence: \texttt{niuyanan@kuaishou.com}.}}
\affil{Kuaishou Technology}
\date{}

\begin{document}
\maketitle

\begin{abstract}
We study factual edit propagation in a controlled synthetic knowledge-graph QA setting using a $2{\times}2$ grid that crosses loop recurrence with shared-memory access: a dense transformer (\runA{}), a looped transformer (\runP{}), a dense backbone with shared memory (\runM{}), and a looped backbone with shared memory (loop--memory coupling, \runD{}). The two factors dissociate. For learning, both routes to repeated shared access---looped recomputation and repeated memory rereading---cross the out-of-distribution (OOD) grokking barrier that \runA{} fails, so repeated shared access is the behavioral regularity, not a specific architecture. For editing, the substrates split along a different axis. Applying a single localized factual edit and measuring propagation on a shared, pre-edit-correct in-distribution set (conditioned on direct edit success), the edit propagates through 2-hop compositions strongly in both memory-bearing cells (\runD{} $0.78$--$0.92$, \runM{} $0.71$--$0.96$) and only weakly in the memory-free ones (\runP{} $0.04$--$0.30$, \runA{} $0.00$--$0.03$). The split is along the memory axis, not the loop axis: we do not detect a difference between the two memory cells at our sample size, while every memory-bearing seed exceeds every memory-free seed. Crucially \runM{} has no recurrence, so the propagating ingredient is an addressable site that an edit can write to and that later computation rereads, not loop recomputation; \runP{}, which reuses a fact-bearing computation but for which we do not identify a separately-writable site under our edit protocol, is at best a partial intermediate case. Within \runD{}, the same edit on \runD{}'s own pre-edit-correct probe set achieves $100\%$ direct success, mean $0.989$ 2-hop propagation, and ${\approx}0.1\%$ movement of random unrelated probes on that internal set (a held-out unrelated-fact set moves ${\approx}1.5\%$; co-resident facts can move more)---a within-\runD{} precision characterization, not a recount of the cross-substrate numbers above. The affordance survives coarsening the store ($N{=}128 \to N{=}13$): propagation attenuates but the memory/no-memory split persists, so fine granularity buys precision rather than the affordance itself. These results dissociate learning competence from editing affordance---repeated shared access suffices to grok, but edit propagation depends on whether the substrate exposes an addressable memory that the forward computation can write to and later reread, an affordance that loop recurrence provides only partially.
\end{abstract}

\section{Introduction}
\label{sec:intro}

\paragraph{Scope.} This paper is a controlled mechanism study---a \emph{model organism} for edit
propagation (a deliberately simplified, fully observable system studied to isolate one mechanism, by
analogy with model organisms in biology), not an architecture proposal. The synthetic KG-QA setting and small five-seed
design are chosen so that the fact-bearing site is causally localizable and the propagation
protocol can be conditioned on direct edit success; we do not claim, and the design does not
test, generalization to natural-language pretraining or to retrieval-augmented (RAG) systems. Within
that scope we adjudicate one question: when two architectures both cross the OOD grokking
barrier, what makes a successful single-fact edit propagate through composition?

Dense transformers can learn the in-distribution surface of a compositional task while failing at the
compositions that matter. In the synthetic knowledge-graph QA setting of \citet{wang2024grokked}, a standard
dense transformer memorizes training and in-distribution (ID) facts but remains near chance on held-out
out-of-distribution (OOD) 2-hop
compositions. Two modifications fix this failure: a looped transformer can recompute with the same backbone
across recurrent depth, and an LMC-style model can reread an explicit memory store between iterations. The
central question of this paper is what these two successful routes have in common, and where they differ.
Throughout, we use \emph{grokking} for the delayed onset of OOD generalization. \citet{wang2024grokked} treat
it as a binary event (a model either does or does not eventually generalize OOD); we adopt that binary view but
make the onset measurable by operationalizing grokking as the first training
step at which OOD accuracy crosses $0.1$ ($\approx25\times$ the chance level), treating $0.1$ as a conservative
above-chance event marker rather than a saturation criterion. We use grokking only as an onset event---\emph{whether} and \emph{when} a configuration crosses this
threshold; the eventual OOD value enters only to confirm that grokked runs clear chance by a wide margin, and
speed claims below refer to time-to-onset.

The two chance levels used below correspond to different metrics: OOD 2-hop accuracy is scored against the
candidate tail-entity set of the KG-QA evaluation (the set of entities admissible as answers, which is far
smaller than the full vocabulary; chance ${\approx}0.004$), while atomic single-hop recall is
scored by exact match over the full word-level vocabulary (chance ${\approx}1/2202\approx5\times10^{-4}$). We
name which baseline applies whenever we refer to chance.

This behavioral regularity is \emph{repeated shared access}. A looped transformer obtains
multiple chances to use the same fact-bearing computation by applying tied backbone weights at several
iterations; a memory-augmented loop obtains multiple chances to query the same fact-bearing store.

This view distinguishes two levels. At the learning level, loop and memory are unified at the level of
crossing the OOD grokking barrier where a non-sharing dense baseline does not (we do not claim
identical training dynamics or representations). At the editing level, however, the relevant axis is
not loop-versus-memory but \emph{addressability}: a substrate that exposes an addressable store (LMC, and also a
dense backbone with shared memory) lets a localized edit propagate, whereas a substrate that shares only
computation does so at best partially. By \textbf{addressable} we mean that each stored fact occupies an
identifiable site that can be written to in isolation and that the forward computation rereads when it needs
the fact. A memory is \textbf{fine-grained} to the degree that this fact-to-site mapping is close to one-to-one,
so editing a single site moves the target fact with little collateral (this is the $N{=}128$ vs.\ $N{=}13$ factor
of \S\ref{sec:discussion}). Loop recurrence reuses a fact-bearing computation across iterations but exposes no
such separately-writable site, which is why it is an intermediate, partial case on the editing axis rather than
a co-equal route.

This distinction frames the paper's central question. If edit propagation follows only the repeated-access
structure, then editing a reused loop site should behave like editing a reused memory site: loop and memory
would be the same all the way down. If, instead, edit propagation depends on explicit addressability and on
when edited content becomes usable, then memory should afford stronger and cleaner propagation than tied looped
weights. We adjudicate between shared-access and addressability accounts rather
than assuming the memory answer in advance.

Our results support a refined version of the addressability account. The final pattern is not a binary
``memory propagates, loop does not.'' Rather, edit propagation splits the substrates along whether they expose
an addressable site: the two memory-bearing configurations propagate strongly with no difference we detect at our sample size,
the dense no-memory model does not propagate, and the looped no-memory model is a partial,
intermediate case. We also explored an interchange diagnostic to ask when edited content becomes usable, but it does
not predict propagation at the seed level (it saturates within LMC and anti-correlates with propagation within
Loop), so we report it only as an auxiliary observation in Appendix~\ref{app:asp-calibration} and do not build
the account on it.

\paragraph{Positioning.} Our task and grokking recipe follow \citet{wang2024grokked}, who predict that
cross-layer memory sharing---via memory augmentation or recurrence---enables OOD systematicity; concurrent
loop-only work \citep{kohli2026loopthink} confirms the recurrence half. We add the controlled loop-vs-memory separation and, distinctively,
the edit-propagation adjudication, holding effective depth fixed in the main grid and using same-scale controls
to separate route viability from parameter count. Full related work is in \S\ref{sec:related}.

\paragraph{Contributions.}
Two terms recur below: a \emph{step} is one loop iteration (step0$\ldots$step3), distinct from a \emph{hop}, which is a reasoning edge; a fact's \emph{site} is the dominant memory value row at the step where the fact is read (full definitions in \S\ref{sec:task}).
\begin{enumerate}[leftmargin=1.6em,itemsep=2pt]
  \item \textbf{Repeated shared access as a behavioral regularity for grokking.} A single-shared-FFN memory
        control and a partial-sharing loop control each grok, while dense no-memory
        models remain flat.
        Behaviorally, crossing the OOD barrier tracks repeated access to a shared substrate rather than sparse
        MoE routing or full recurrence specifically; we make no claim that the two routes share an internal
        circuit.
  \item \textbf{Localization from grokking to editable addressability.} In the fine-grained LMC memory, causal
        patching and interchange interventions show that one dominant memory site stores each fact and is reused
        in 2-hop reasoning: the
        single-hop step0 dominant expert equals the 2-hop step1 dominant expert for every tested atom on each of the five LMC seeds (selection protocol in Appendix~\ref{app:localization-protocol}),
        and bridge substitution flips the 2-hop answer in 92--99.5\% of cases across all five LMC seeds; the
        symmetric coarse interchange diagnostic (one site per loop iteration) returns best-site flip-to-donor of $1.00$ on every
        seed (Section~\ref{sec:bridge-results}, Appendix~\ref{app:localization-perseed}).
  \item \textbf{Local LMC editing.} Editing only the largest activated value row at
        the step0 edit site gives 100\% direct success (LMC-internal probe set) and mean 2-hop bridge propagation
        $0.989$ on that set; on the cross-substrate shared ID measurement set (Contribution~4),
        the same edit propagates at $0.78$--$0.92$ across seeds.
        Locality has three distinct scopes: random unrelated probes move ${\approx}0.1\%$ (internal set, Table~\ref{tab:lmc-edit}),
        a matched held-out unrelated-fact set moves ${\approx}1.5\%$ (Table~\ref{tab:specificity}), and
        co-resident facts sharing the edited expert can move much more (5--50\%, Appendix~\ref{app:coresident}).
        Locality is therefore not absolute: hidden-state overlap predicts the co-resident leakage.
  \item \textbf{A 2$\times$2 edit-propagation adjudication.} Using a shared ID measurement set,
        pre-edit-correct probes, and conditioning on direct edit success, we compare \runA{}, \runP{}, \runM{}
        (a dense backbone with the same memory store but no loop), and \runD{}. The result ties high semantic
        propagation to memory-bearing cells (\runD{} $0.78$--$0.92$,
        \runM{} $0.71$--$0.96$), while \runP{} remains intermediate ($0.04$--$0.30$) and \runA{} remains near zero ($0.00$--$0.03$).
\end{enumerate}

The rest of the paper follows this arc: we first define the KG-QA setup and architecture grid, then present
Study~1 on learning, a localization bridge that establishes editable memory reuse, and Study~2 on edit
propagation. We return to related work after the main evidence, where the contrasts are easier to evaluate.

\section{Experimental Setup and Protocols}
\label{sec:method}

This section defines the task, the architectural variants, and the intervention protocol used in the
edit-propagation adjudication. The central design discipline is that learning and editing are evaluated with
different controls: Study~1 asks which repeated-access structures cross the OOD grokking barrier, whereas
Study~2 asks whether a successful direct edit propagates through a composition.

\subsection{Task: synthetic knowledge-graph QA}
\label{sec:task}

We use the synthetic knowledge-graph QA setting of \citet{wang2024grokked}. The data are generated from a
set of atomic facts $(e,r)\mapsto e'$ and composed 2-hop queries of the form
$(e_0,r_1,r_2)\mapsto e_2$, where $e_0 \xrightarrow{r_1} e_1 \xrightarrow{r_2} e_2$ and the intermediate entity
$e_1$ is the \emph{bridge}. Models are trained on
single-hop and in-distribution examples and evaluated on held-out OOD compositions. Recall from
\S\ref{sec:intro} that we operationalize grokking as the first step at which OOD accuracy exceeds $0.1$, and
use only \emph{whether} and \emph{how fast} a configuration crosses it; we do not rank substrates by the
eventual OOD value.

The task is useful for the present question because it separates memorizing atomic or ID facts from
systematically composing facts. Dense no-memory transformers can fit training and ID data while remaining
near random on held-out OOD 2-hop compositions; repeated shared access through a loop or a memory can cross
that barrier.

\paragraph{Notation and terminology.}
We collect the recurring terms in one place so that later sections can use them without repeated
re-introduction.

\emph{Step vs. hop.} A step is one loop iteration of the shared backbone. With $R{=}4$ we index the
iterations as \emph{step0\,$\ldots$\,step3}, read memory after step0/1/2, and leave step3 as a
memory-free finishing pass before the LM head (see \S\ref{sec:variants}). The choice $R{=}4$ is fixed by
depth-matching rather than tuned: $3\text{L}\times4$ gives the same $12$ effective layers as the dense and
partial-sharing baselines, and four iterations suffice to run a 2-hop chain followed by a memory-free finishing
step. A \emph{hop} is a reasoning edge in the query and is not a step.

\emph{The 2-hop query.} For $(e_0,r_1,r_2)\mapsto e_2$ with
$e_0 \xrightarrow{r_1} e_1 \xrightarrow{r_2} e_2$, we call $e_1$ the \emph{bridge entity},
$(e_0,r_1)\mapsto e_1$ the \emph{first-hop fact}, and $(e_1,r_2)\mapsto e_2$ the
\emph{second-hop fact}. The computation is iterative single-hop: step0 resolves the first-hop fact and emits
the bridge entity; step1 rereads that bridge entity and resolves the second-hop fact; step2 produces the answer
representation. Section~\ref{sec:bridge-results} measures the same-location reuse that makes a single memory
edit meaningful: the second-hop fact uses the same dominant site whether queried directly at step0 or reached
compositionally at step1.

\emph{Site, edit, success, propagation.} A fact's \emph{site} is the dominant value row of its dominant memory
expert at the step where the fact is read. Facts that share the same dominant memory expert are
\emph{co-resident facts}. An edit retargets one fact at its site. In the propagation study we edit the
first-hop fact, replacing its bridge entity $e_1$ with $e_1'$ (chosen so that $(e_1',r_2)$ is an existing fact).
\emph{Direct edit success} means $(e_0,r_1)$ now predicts $e_1'$. Conditioned on direct success,
\emph{strong propagation} means the 2-hop prediction equals the answer implied by the new bridge entity,
$\mathrm{fact}(e_1',r_2)=e_2'$, whereas \emph{weak propagation} is any change away from the original answer
$e_2$; the main text uses strong unless noted.

\emph{Substrate split.} A substrate is \emph{memory-bearing} if it has a shared memory store---LMC and the
dense-with-memory control---and \emph{memory-free} otherwise (Loop and Dense). The adjudication asks whether
edit propagation tracks the memory-bearing/memory-free split or the backbone (looped/dense) split.

\subsection{Architectural variants}
\label{sec:variants}

We organize the main KG-QA architectures as a 2$\times$2 control grid (Table~\ref{tab:config-grid}) that crosses backbone type with shared-memory
access.
\begin{table}[tbp]
\centering
\caption{Primary configuration grid. \runA{} is the dense no-memory anchor, \runP{} isolates repeated
recomputation, \runM{} is the dense-backbone, memory-only control, and \runD{} ($\star$) is the full model: looped
backbone plus shared memory. Horizontal arrows add shared memory; vertical arrows add loop recurrence.}
\label{tab:config-grid}
\begin{tikzpicture}[
    font=\small,
    cell/.style={draw,rounded corners,thick,text width=34mm,minimum height=13mm,align=center,inner sep=2pt},
    anchor cell/.style={cell,draw=gray!60,fill=gray!8},
    mem cell/.style={cell,draw=blue!50!black,fill=blue!7},
    lmc cell/.style={cell,draw=red!60!black,fill=red!8},
    axlbl/.style={font=\footnotesize\bfseries},
  ]
  \node[axlbl] at (0,1.35)    {no memory};
  \node[axlbl] at (4.2,1.35)  {shared memory};
  \node[axlbl,rotate=90] at (-2.6,0.0)   {dense};
  \node[axlbl,rotate=90] at (-2.6,-1.7)  {looped};
  \node[anchor cell] (a) at (0,0.0)    {\runA{}\\\scriptsize neither (anchor)};
  \node[mem cell]    (m) at (4.2,0.0)  {\runM{}\\\scriptsize memory alone};
  \node[mem cell] (p) at (0,-1.7)    {\runP{}\\\scriptsize recurrence alone};
  \node[lmc cell] (d) at (4.2,-1.7)  {\runD{}$\,\star$\\\scriptsize recurrence $+$ memory};
  \draw[->,thick,orange!75!black] (1.55,1.35) -- (2.65,1.35) node[midway,above=1pt,font=\scriptsize]{$+$mem};
  \draw[->,thick,blue!55!black]   (-2.6,-0.6) -- (-2.6,-1.05) node[midway,left=2pt,font=\scriptsize]{$+$loop};
\end{tikzpicture}
\end{table}
The grid separates two factors---repeated computation through a looped backbone, and repeated access to a shared
memory. The edit-propagation comparison uses all four cells: \runA{} tests dense no-memory editing, \runP{}
tests loop-only editing, \runM{} tests memory without loop recurrence, and \runD{} combines loop and memory.

Model scale is reported as an architectural covariate rather than force-matched, because the editing protocol
aligns on the fact-bearing site and on direct edit success rather than on parameter count (we separately add
parameter-matched same-scale controls in \S\ref{sec:axis2runs}, ruling out a parameter-count explanation). Scale differences do
not explain the propagation gap: every substrate reaches a successful direct edit (Loop and Dense at 100\%
direct success on all five seeds), so the differences arise after the edit takes, in whether it propagates, not
in editability or in raw capacity.

\begin{figure}[tbp]
  \centering
  \resizebox{\linewidth}{!}{%
  \begin{tikzpicture}[
      font=\normalsize,
      lyr/.style={draw=green!45!black,thick,rounded corners,fill=green!15,
                  minimum width=14mm,minimum height=5.5mm,align=center,font=\footnotesize},
      side/.style={draw,rounded corners,thick,minimum width=14mm,minimum height=9mm,align=center},
      emb/.style={side,draw=blue!55!black,fill=blue!10},
      ln/.style={side,draw=red!55!black,fill=red!10},
      mem/.style={draw=orange!70!black,very thick,rounded corners,fill=orange!12,
                  minimum height=11mm,align=center},
      stack/.style={draw=green!45!black,thick,dotted,rounded corners,fill=green!4,inner sep=2mm},
      io/.style={align=center,font=\footnotesize},
      fwd/.style={->,thick,>={Stealth[length=2.3mm]}},
      rd/.style={->,thick,orange!85!black,>={Stealth[length=2mm]}},
      hlbl/.style={font=\footnotesize,above,inner sep=1pt},
    ]
    \node[io] (in) {input\\\footnotesize tokens};
    \node[emb,right=8mm of in] (emb) {Embedding};

    \node[lyr,right=12mm of emb] (a2) {Layer 2};
    \node[lyr,above=2mm of a2] (a1) {Layer 1};
    \node[lyr,below=2mm of a2] (a3) {Layer 3};

    \node[lyr,right=16mm of a2] (b2l) {Layer 2};
    \node[lyr,above=2mm of b2l] (b1l) {Layer 1};
    \node[lyr,below=2mm of b2l] (b3l) {Layer 3};

    \node[right=7mm of b2l,font=\large] (dots) {$\cdots$};

    \node[lyr,right=7mm of dots] (r2l) {Layer 2};
    \node[lyr,above=2mm of r2l] (r1l) {Layer 1};
    \node[lyr,below=2mm of r2l] (r3l) {Layer 3};

    \begin{scope}[on background layer]
      \node[stack,fit=(a1)(a3)] (s1) {};
      \node[stack,fit=(b1l)(b3l)] (s2) {};
      \node[stack,fit=(r1l)(r3l)] (sR) {};
    \end{scope}
    \node[font=\footnotesize,green!45!black,above=0.5mm of s1.north] {step $0$};
    \node[font=\footnotesize,green!45!black,above=0.5mm of s2.north] {step $1$};
    \node[font=\footnotesize,green!45!black,above=0.5mm of sR.north] {step $R{-}1$};

    \node[ln,right=12mm of sR] (ln) {Layer\\Norm};
    \node[emb,right=7mm of ln] (head) {LM\\Head};

    \foreach \t/\m/\b in {a1/a2/a3, b1l/b2l/b3l, r1l/r2l/r3l}{
      \draw[->,thin,green!45!black] (\t) -- (\m);
      \draw[->,thin,green!45!black] (\m) -- (\b);
    }

    \draw[fwd] (in) -- (emb);
    \draw[fwd] (emb) -- node[hlbl]{$x^{(0)}$} (s1.west|-emb);
    \draw[fwd] (s1.east|-a2) -- node[hlbl]{$x^{(1)}$} (s2.west|-b2l);
    \draw[fwd] (s2.east|-b2l) -- node[hlbl]{$x^{(2)}$} (dots);
    \draw[fwd] (dots) -- node[hlbl]{$x^{(R-1)}$} (sR.west|-r2l);
    \draw[fwd] (sR.east|-r2l) -- node[hlbl]{$x^{(R)}$} (ln);
    \draw[fwd] (ln) -- (head);

    \draw[decorate,decoration={brace,amplitude=4pt,mirror},gray!70]
      ([yshift=-3mm]s1.south west) -- ([yshift=-3mm]sR.south east)
      node[midway,below=3pt,font=\footnotesize,gray!70!black]
        {shared weights $f_\theta$, looped $R$ times};

    \node[mem,minimum width=78mm] at ($(s1.south)!0.5!(sR.south)+(0,-15mm)$) (mem)
      {\textbf{shared memory} $\mathrm{Mem}_\phi$};
    \foreach \s in {s1, s2}{
      \draw[rd] (\s.south |- mem.north) -- (\s.south);
    }
    \draw[rd,dashed,opacity=0.45] (dots.south |- mem.north) -- (dots.south);
    \node[font=\footnotesize,orange!60!black,below=1.5mm of mem.south]
      {queried each \emph{non-final} iteration};
  \end{tikzpicture}%
  }
  \caption{Loop-Memory Coupling architecture. A thin shared backbone ($L{=}3$ layers) is iterated $R{=}4$ times
  with tied weights $f_\theta$, giving $12$ effective layers. At each non-final iteration the latent state queries
  the same shared memory $\mathrm{Mem}_\phi$ and adds the retrieval by a residual connection; the final iteration
  is a pure-transformer finish before the LM head. Removing the memory recovers the pure looped backbone (\runP);
  replacing the loop with a single-pass $12$-layer backbone that rereads the same memory at three matched access
  points recovers the dense-with-memory cell (\runM); removing both recovers the dense anchor (\runA).
  Iterations are indexed step0$\ldots$step$R{-}1$ (the first iteration is step0), matching the body notation; the
  shared memory is read after step0, step1, step2, and the final iteration (step3) is memory-free.}
  \label{fig:arch}
\end{figure}
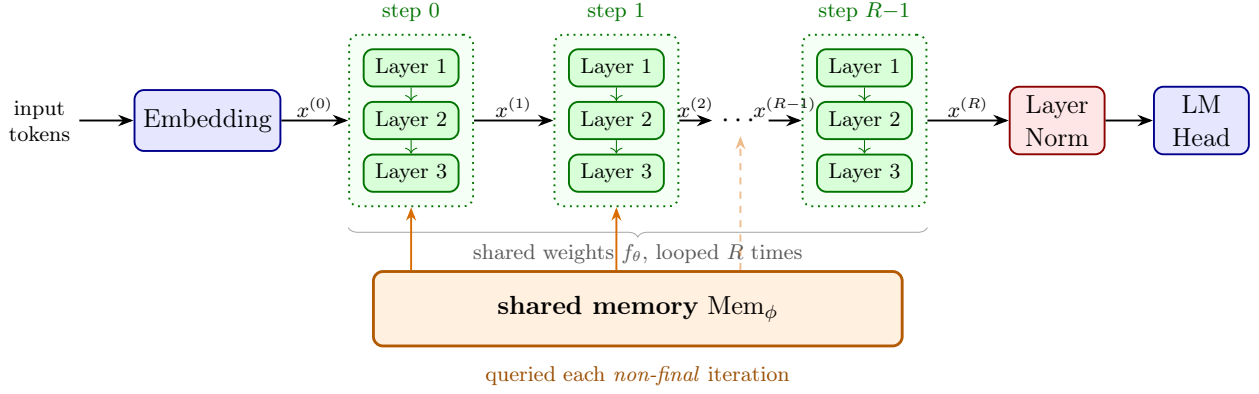

\paragraph{Loop-memory coupling.}
LMC consists of a shared transformer block $f_\theta$ of $L$ layers together with a shared memory
$\mathrm{Mem}_\phi$ (Figure~\ref{fig:arch}). Given token embeddings $x^{(0)}$, the model iterates the shared
block $R$ times and queries memory between iterations:
\begin{align}
  h^{(r)} &= f_\theta(x^{(r)}), \\
  x^{(r+1)} &=
  \begin{cases}
    h^{(r)} + \mathrm{Mem}_\phi(\mathrm{LN}(h^{(r)})), & r < R-1, \\
    h^{(r)}, & r = R-1 .
  \end{cases}
\end{align}
The memory is implemented as a sparsely routed expert store: a router selects top-$k$ expert MLPs and mixes
their outputs. The paper's claims do not depend on sparse routing alone; Study~1 explicitly tests a single
shared-FFN memory variant.

\paragraph{Memory placement across the variants.}
The same parameters $\theta,\phi$ are reused across all $R$ iterations; the only per-step variation comes from
the evolving latent $h^{(r)}$ and its memory retrieval, and we train end-to-end with the next-token loss using
\emph{no} step-level supervision and \emph{no} curriculum. For the dense-with-memory control (\runM{}), the backbone is a single-pass dense transformer rather than a loop,
but the memory principle is kept identical: one shared memory is reread at each configured access point. In the $12$-layer \runM{} cell the shared
memory is queried after layers $3$, $6$, and $9$, matching the three non-final reads of LMC, while the terminal
layer stays memory-free so the final retrieval is processed by later layers before the LM head. Removing the
memory from LMC recovers a pure looped transformer (\runP), and removing both recurrence and memory recovers the
dense baseline (\runA); together with full LMC (\runD) these form the $2\times2$ grid above.

\subsection{Study 1 protocol: repeated shared access}
\label{sec:study1protocol}

Study~1 tests whether OOD grokking tracks repeated access to a shared substrate rather than an architecture
label. We use two diagnostics.

\paragraph{Single-shared-FFN memory control.}
The MoE memory is replaced by one dense shared FFN inserted at the same repeated access points. If this model
groks, sparse MoE routing is not necessary for the existence of grokking under the recipe.

\paragraph{Partial-sharing loop control.}
We train a partially shared looped model with $4$ unique layers applied $3$ times (4L$\times$3, still $12$
effective layers), so its degree of weight tying lies between the dense $12$-layer baseline (no sharing) and the
full 3L$\times$4 loop (maximal sharing). This tests whether the full recurrence pattern is necessary, or whether
repeated shared access at an intermediate degree of sharing also crosses the OOD barrier.

Both diagnostics are reported across five seeds. Auxiliary controls support the same conclusion: dense no-memory
stays flat, one memory access is insufficient, and at least two memory accesses can cross the OOD barrier.

\subsection{Localization measurements: addressability and iterative single-hop reuse}
\label{sec:bridgeprotocol}

The localization step from grokking to editing is the claim that LMC facts are both localized and reused by composition.
We use three intervention measurements, with the step/hop terminology of \S\ref{sec:task}: editing at step0
means writing to the memory site that reads the first-hop fact.

\paragraph{Operational single-store localization.}
For each atomic fact, we identify the dominant memory expert at the direct single-hop read step. Fine-grained
memory yields an operational single-store regime: within the measured memory read, one dominant expert/value
site causally controls the measured fact and is reused in 2-hop. We do not claim that no other parameter
carries correlated information. The dominant expert carries weight $0.93$--$0.98$ across seeds, and
wrong-expert interventions leave recall intact.

\paragraph{Same-location reuse.}
We compare the expert used by the direct single-hop fact with the expert used when the same fact appears as
the second-hop fact in a 2-hop query. The key measurement is whether the single-hop step0 dominant
expert equals the 2-hop step1 dominant expert. This is the precondition for interpreting a single memory edit
as an edit to the location composition will later read.

\paragraph{Interchange connectivity.}
We also use bridge substitution: replace the answer-position memory input at the 2-hop step with the
corresponding vector from a donor query sharing $r_2$ but having a different bridge entity. If the answer flips to
the donor answer, the bridge-entity information in the residual/memory input is causally controlling the 2-hop
answer. This distinguishes iterative single-hop composition from a direct memory lookup of a whole
2-hop composition.

\subsection{Study 2 protocol: edit propagation}
\label{sec:editprotocol}

Study~2 uses one common edit-propagation protocol across LMC, Loop, Dense+Mem, and Dense. For each substrate, we
first localize the smallest causally effective site for the first-hop fact, apply a ROME-style value write
\citep{meng2022rome}
toward a replacement bridge entity $e_1'$, and sweep the edit budget until the direct atomic answer changes. The exact
site definitions and sparse updates are in Appendix~\ref{app:edit-protocol}.

The edit comparison is restricted to the $2\times2$ grid because that grid is what isolates the two editing-relevant
factors---backbone recurrence and shared-memory access. The Study~1 auxiliary controls (the single-shared-FFN
memory and the partial-sharing loop) vary only the \emph{internal implementation} of an already-present factor
(memory granularity and degree of weight tying); they were designed to test the necessity of sparse routing and
full recurrence for \emph{learning}, and add no new addressability axis for editing. In particular, the
single-shared-FFN control is, for editing purposes, on the same memory-bearing side as LMC and Dense+Mem, so we
do not treat it as a separate edit-propagation condition.

Propagation is measured only on compositions the model answered correctly before the edit and only after direct
edit success. A strong propagation event requires the edited 2-hop answer to equal the answer implied by the new
bridge,
\begin{equation}
  \hat{y}_{\text{2-hop after edit}} = \mathrm{fact}(e_1', r_2),
\end{equation}
whereas a weak event records any change away from the original answer. The main text uses the strong metric;
weak propagation is reported as a secondary diagnostic in Appendix~\ref{app:weak-prop}.

To interpret timing, we use an interchange diagnostic that swaps bridge information while holding the final-answer
direction fixed. A flip under this answer-subspace-preserving interchange is consistent with usable bridge information being present,
not with direct answer copying. The formal definition and calibration are in Appendix~\ref{app:asp-def}.

\section{Study 1: Repeated Shared Access Enables OOD Grokking}
\label{sec:study1}

Study~1 shows that OOD grokking follows repeated shared access rather than a specific implementation.

\begin{table}[t]
\centering
\caption{Study~1 grokking summary. All configurations use 12 effective layers. The grok step is the first
training step at which OOD accuracy exceeds 0.1; per-seed values and full configuration details are in
Appendix~\ref{app:study1-perseed} and Appendix~\ref{app:config}.}
\label{tab:study1-grok}
\small
\begin{tabular}{llcc}
\toprule
Model & Shared-access route & Seeds grokked & Grok step \\
\midrule
Dense & none & 0/5 & never \\
Shared-FFN memory control & repeated memory read & 5/5 & 80k--105k \\
Partial-sharing loop control & partial recurrence & 5/5 & 220k--230k \\
Loop 3L$\times$4 & tied-backbone recurrence & 5/5 & 230k--245k \\
Dense+Mem (12L dense) & repeated memory read & 5/5 & 60k--85k \\
LMC $N=128$ & routed memory read & 5/5 & 45k--65k \\
\bottomrule
\end{tabular}
\end{table}

\begin{figure}[t]
\centering
\includegraphics[width=0.92\linewidth]{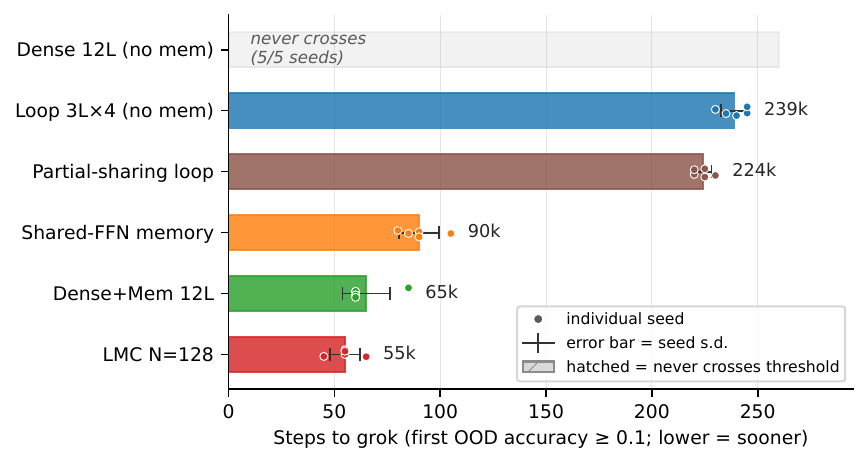}
\caption{\textbf{OOD grokking as a binary onset event.} Following the binary view of grokking in \citet{wang2024grokked}, we plot only whether each configuration crosses the conservative above-chance threshold and, when it does, the first step at which held-out two-hop OOD accuracy reaches $0.1$. Bars show five-seed mean onset steps; dots show individual seeds; error bars show seed standard deviations. The Dense 12L anchor never crosses on any seed and is shown as a hatched ``never'' bar. These are the primary, non-parameter-matched cells, so onset steps are descriptive only and are not efficiency or speed comparisons across cells; parameter-matched (Iso) onsets are reported separately in \S\ref{sec:axis2runs}. The figure intentionally does not compare final OOD accuracy values or saturation levels; speed claims in Study~1 refer to time-to-onset only.}
\label{fig:grok-onset-steps}
\end{figure}

The controls separate repeated access from implementation details. The shared-FFN memory control removes sparse
MoE routing but keeps repeated memory reads, and it groks on every seed. The partial-sharing loop control reduces
the degree of weight tying, and it also groks. The dense-backbone memory-only cell (\runM{}, a dense backbone
with the same memory store but no loop recurrence) also groks (60k--85k steps), confirming that repeated memory reads cross the barrier even without loop recurrence;
this is the same cell used as the memory-only control in the Study~2 edit comparison. Dense no-memory remains
flat. These results support the repeated-shared-access account; per-seed grok steps are in
Table~\ref{tab:app-perseed-grok}.

Memory also changes the quantitative regime in the four-cell grid. The fine-grained $N{=}128$ LMC run reaches grok onset (OOD$>0.1$)
in 45k--65k steps, about 4.4$\times$ fewer than the loop baseline.\footnote{Per-step compute differs across cells (Loop reuses 3 unique layers $R{=}4$ times; LMC additionally queries a 128-expert memory store; Dense uses 12 unique layers), so this is a step-count comparison, not an iso-FLOP efficiency claim.} This four-cell speedup is largely a
capacity effect, not an advantage of memory per se. Once parameter count is matched (\S\ref{sec:axis2runs}), the ordering reverses: a
loop-only model (\runPbig{}, ${\approx}146$k steps) reaches onset \emph{faster} than the same-scale
dense-memory model (\runMiso{}, ${\approx}449$k), with the same-scale LMC (\runDiso{}, ${\approx}166$k)
comparable to \runPbig{}. The same-scale runs use a larger configuration than the four-cell models above, so absolute step counts are not directly comparable across the
two tables; Table~\ref{tab:app-config-provenance} disambiguates the four-cell \runD{} from the same-scale
\runDiso{}. Routing
granularity is not required for crossing the
barrier, but it sharpens atom-to-site addressability for the editing study; all editing results below are at
this fine granularity, and we return to the resulting granularity confound as a scope limit in
Section~\ref{sec:discussion}.

\subsection{Auxiliary route-viability controls (same-scale and access-count)}
\label{sec:axis2runs}
The four-cell $2{\times}2$ design above varies recurrence and shared memory along the axes that
matter for the editing study, but two further controls are useful for separating ``repeated shared
access'' from confounds (parameter count, mere presence of a store). We collect them here.

\paragraph{Same-scale route comparison.}
The four-cell grid does not match parameter counts across cells. We therefore add a same-scale
control pair at the same $12$ effective layers and the same $\approx87$M-class budget (the ``-Iso'' suffix denotes this scale-matched setting): \runPbig{}
(widened loop-only backbone) versus \runMiso{} (narrowed dense backbone with shared memory). \runDiso{}
is the mixed reference. Both controls cross $0.1$ on OOD, with the loop route faster
($146{,}000\pm6{,}500$ vs.\ $449{,}000\pm134{,}100$ steps); \runDiso{} sits close to \runPbig{} at
$166{,}000\pm26{,}600$. The memory-only route is far more variable: every \runPbig{} seed groks
before every \runMiso{} seed ($140$--$155\text{k}$ vs.\ $310$--$640\text{k}$), maximal rank
separation (Mann--Whitney $U{=}25/25$, exact two-sided $p{=}0.008$). This is a route-viability
contrast, not a parameter-efficiency claim: with depth and scale fixed, recomputation and rereading
both produce the OOD transition, and the pure loop route gets there sooner. Both routes are thus
viable for learning---each crosses the barrier that Dense never does---but they are not symmetric: at
matched scale the loop route is significantly faster to onset (complete rank separation, $p{=}0.008$).
The asymmetry that the editing study turns on runs the other way (memory, not loop), so the two
axes are each asymmetric but in opposite directions (\S\ref{sec:study2}).

\paragraph{Access-count ablation suggests a repeated-read threshold.}
Holding the looped backbone and parameters fixed and varying only how many times the shared memory
is consulted per forward pass, the access-count controls show a sharp qualitative split: a single
access (read-once; $n{=}5$, including a reproduction of \citet{wang2024grokked}'s one-access
negative control) does not grok, whereas two or more accesses do (twice at $165\text{k}$, $n{=}5$;
every-iteration---i.e.\ \runDiso{}---at $166\text{k}$, $n{=}5$). The boundary therefore lies between
one and two reads; beyond two, the current estimates are close ($165\text{k}$ vs.\ $166\text{k}$).
The operative variable for the memory route in these controls is repeated access, not the mere
existence of a store, supporting the unifying interpretation that repetition---whether by
re-reading or by re-computing---is a common operational ingredient behind the observed transition.
This is a behavioral claim about interventions, not evidence that the two routes implement the same
internal circuit.

\subsection{Training without step-level supervision (supporting control)}
\label{sec:noaux}
For the shallow recurrent backbone to \emph{compute} across iterations rather than merely refine a fixed
answer, the intermediate steps must be allowed to diverge from the final output---a precondition for the
whole setup, which we verify here.
Throughout, all configurations are trained with \emph{no} step-level auxiliary loss. We find this matters: a
step-level shared-target auxiliary next-token loss collapses the recurrence into iterative refinement (cosine
similarity between first- and last-iteration latents rises from $0.47$ to $0.82$). This refines,
rather than contradicts, prior work arguing that latent reasoning benefits from step-level
supervision \citep{simcot}: the operative variable is the \emph{diversity} of the supervision
target, not its presence (Appendix~\ref{app:auxloss}).

\section{Behavioral Localization: Recall and Composition Have Different Damage Thresholds}
\label{sec:localization}
After Study~1 establishes that LMC and Dense+Mem cross the OOD grokking barrier, the remaining question before
the editing study is mechanistic: when the memory route matters, does it merely add raw capacity, or does it
create a separable route for storing, addressing, and using facts? We answer this with destructive interventions
on trained \runD{} checkpoints. The resulting claim is behavioral rather than symbolic: factual recall,
memory addressability, and compositional use have different damage thresholds under intervention.
We characterize the addressable site in \runD{} because it is the finer-grained of the two memory cells;
the cross-substrate adjudication (\S\ref{sec:three-substrate-results}) then shows the same affordance in
\runM{}, which has no loop, so the site---not the loop---is what carries propagation.

\paragraph{Frozen-recall knockout.}
We use a causal-localization test in the spirit of \citet{meng2022rome}: take the trained \runD{} model, and at
inference \emph{ablate the external memory} (zero its contribution, leaving the backbone untouched), then measure
how much single-hop factual recall survives. The probe set is the atomic facts themselves---every single-hop
$(\text{head},\text{rel})\!\to\!\text{tail}$ triple, each of which appears verbatim in the training data---so the
test isolates stored \emph{knowledge} from multi-hop reasoning. Our decision variable is the
\emph{within-model} drop in recall when memory is knocked out; a memory-free looped backbone (\runP{}),
trained without any store, serves as a context anchor (the recall a same-family backbone reaches when it
\emph{must} store facts internally). We evaluate a fixed random probe subset of $4{,}000$ atomic facts (of the ${\approx}40{,}000$ in the KG; see Table~\ref{tab:app-training}) under greedy exact-match.

\begin{table}[t]
  \centering
  \caption{Frozen-recall knockout on the atomic single-hop facts ($N{=}4{,}000$, greedy exact-match;
  means over $n{=}5$ seeds). Ablating the external memory drops the LMC model's factual recall
  from near-perfect ($0.998$) to a substantial-but-minority residual ($0.33\pm0.12$), while the same
  backbone left intact in the memory-free anchor \runP{} recalls almost everything---the bulk
  of recall is memory-dependent, with a non-trivial minority surviving in the backbone.}
  \label{tab:frozen}
  \begin{tabular}{lcc}
    \toprule
    Condition & Memory & Atomic recall \\
    \midrule
    \runD{} (full)        & on  & 0.998 \\
    \runD{} (memory off)  & \textbf{off} & \textbf{0.33} \\
    \runP{} (anchor)      & ---{} (none) & 0.998 \\
    \bottomrule
  \end{tabular}
\end{table}

The result is strong but not total (Table~\ref{tab:frozen}). With memory intact, \runD{} recalls $99.8\%$ of
atomic facts; knocking the memory out drops recall to $33\%$ (memory-off recall $0.33\pm0.12$,
mean$\pm$std over $5$ seeds)---a within-model collapse of $0.66\pm0.12$, consistent in direction
across all $5$ seeds. This residual sits \emph{well above} the exact-match chance baseline of
${\approx}5\times10^{-4}$ defined in \S\ref{sec:task}:
the backbone on its own still answers a substantial minority of atomic queries, while the majority of recall is
memory-dependent. The memory-free anchor \runP{}, which had no store to lean on and therefore had to encode
facts in its weights, recalls $99.8\%$. Two readings follow. First, the localization is clear though
not total: the bulk of the model's factual recall depends on the external
memory, while a non-trivial minority is retained in the recurrent backbone. Second, the knockout
leaves the backbone recalling less than the memory-free \runP{} ($0.33$ vs.\ $0.998$). This is the
expected signature of decoupling: \runD{}'s backbone was trained \emph{with} a store and offloaded
fact storage to it, whereas \runP{} had to keep facts in-backbone. We therefore read the verdict
from the within-model drop rather than the \runP{} comparison.

\begin{figure}[t]
  \centering
  \includegraphics[width=\linewidth]{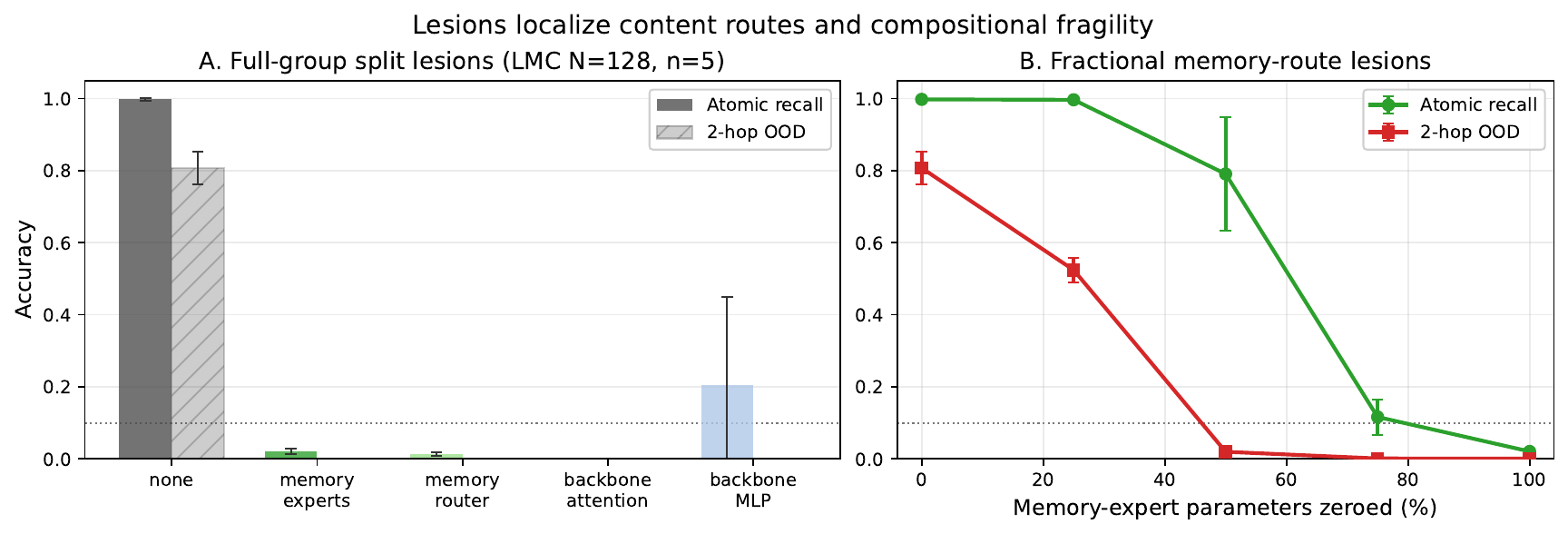}
  \caption{\textbf{Lesions separate factual recall from compositional use.}
  \textbf{A.} Full-group split lesions on \runD{} ($n{=}5$ seeds): zeroing either memory content
  (experts), memory addressability (router), or backbone computation (attention/MLP) drives 2-hop OOD
  to near zero, while atomic recall retains different residuals across groups. \textbf{B.}
  Fractional memory-expert lesions across five checkpoints and three random masks per fraction show a sharper
  threshold difference: $25\%$ expert deletion leaves atomic recall nearly intact but substantially reduces
  2-hop OOD; by $50\%$, OOD is nearly eliminated while atomic recall remains above chance, and the $100\%$
  endpoint matches the full expert-deletion bar in Panel~A.}
  \label{fig:mechanism-lesions}
\end{figure}

\paragraph{Within-LMC split lesions.}
The frozen knockout localizes recall dependence to the store as a whole. To ask whether this is merely
raw capacity or a more structured memory route, we further split trained \runD{} checkpoints into
their memory experts, memory router, and looped backbone components, zeroing each group at evaluation
time while measuring both atomic recall and held-out 2-hop OOD accuracy. Mechanically, this experts-off lesion
differs from the frozen knockout above: the knockout zeros the full memory contribution after the module,
whereas experts-off zeros the expert transformations while leaving the routing interface present.
Panel~A of Figure~\ref{fig:mechanism-lesions} summarizes the five-seed full-group lesions on the
main \runD{} checkpoints ($N{=}128$ experts, top-$k{=}2$, expert hidden $320$; the same
checkpoints used elsewhere in the paper). The unlesioned
checkpoints retain near-perfect atomic recall ($0.998\pm0.004$) and non-trivial 2-hop OOD accuracy
($0.807\pm0.044$). Zeroing the memory experts collapses both behaviours: only $0.020\pm0.008$ atomic
recall and $0.001\pm0.001$ OOD accuracy survive. Zeroing the routing interface ($98{,}304$ parameters,
$768{\times}128$, ${\approx}0.15\%$ of the memory store) is essentially as damaging
($0.013\pm0.005$ atomic, $0.001\pm0.000$ OOD), so addressability is not a negligible detail.
Conversely, zeroing the backbone MLP leaves a small and seed-variable atomic residual
($0.206\pm0.243$, ranging from $0.009$ to $0.624$ across the five seeds) but eliminates OOD
generalization ($0.001\pm0.001$), and zeroing attention destroys both.

The split lesions separate stored content from addressability. Zeroing either the expert transformations or
the routing interface drives atomic recall and 2-hop OOD accuracy near zero, so facts may reside in the
experts but become inaccessible without the address path that selects them. This split-lesion result is sharper
than the frozen memory-off test above: the frozen knockout removes the module output after the memory, whereas
the experts-off lesion zeros the expert transformations themselves while leaving the routing interface present.
We therefore use the frozen knockout only to show that recall is memory-dependent, not to numerically identify
the same residual as the experts-off lesion. Fractional lesions sharpen the same point (Panel~B of
Figure~\ref{fig:mechanism-lesions}).
Across the same five \runD{} checkpoints and three random masks per fraction, deleting $25\%$ of memory-expert parameters
leaves atomic recall essentially intact ($0.997\pm0.004$) but already lowers 2-hop OOD from $0.807\pm0.044$ to
$0.524\pm0.034$; at $50\%$ deletion, OOD has collapsed ($0.020\pm0.008$) while atomic recall is still substantial
($0.791\pm0.157$). At $75\%$ atomic recall has dropped to $0.116\pm0.049$ and OOD remains near zero
($0.001\pm0.001$); the deterministic $100\%$ endpoint coincides with full memory-expert deletion in Panel~A.
Atomic factual recall is therefore strictly more robust to partial memory-route damage than compositional use of
those facts: there is a wide intermediate fraction at which facts can still be \emph{retrieved} but no longer
\emph{composed}.

\paragraph{Dense random deletion as a sensitivity check.}
A matched raw-deletion control on dense no-memory checkpoints is less diagnostic, because deleting random
transformer-block parameters quickly becomes a computation-graph damage test rather than a targeted memory-route
intervention. We therefore report it only as a sensitivity analysis in Appendix~\ref{app:dense-sensitivity}.

\paragraph{Scope.}
These destructive behavioral interventions do not prove a clean symbolic module boundary or quantify bits per
parameter; they establish the substrate used below: atomic facts behave as if they live primarily in LMC memory
experts and are addressed through a lesion-sensitive routing interface.

\section{Mechanism Bridge: LMC Reuses a Dominant Causal Memory Site}
\label{sec:bridge-results}

The transition from Study~1 to Study~2 is the localization of atomic facts in the memory and their reuse during
composition. Three measurements, each reported on all five LMC seeds (s0--s4), support this bridge;
per-seed numbers are in Appendix~\ref{app:localization-perseed}.

\paragraph{Dominant causal site.}
At the direct single-hop step, each tested fact is dominated by one expert site. Across all five seeds, the
step0 dominant routing weight lies in the $0.93$--$0.98$ range; other experts behave like near-tie routing
alternatives. Causal
patching confirms that this is not merely correlational: suppressing the detected primary expert collapses
atomic recall to a mean of $0.169$ across seeds (range $0.020$--$0.460$ when suppressed at every step;
range $0.380$--$0.939$ when suppressed only at step0), while suppressing wrong or random experts leaves
recall essentially intact (mean $1.000$ for wrong-expert, $0.988$ for random-site controls; per-seed
breakdown in Table~\ref{tab:routing-perseed}). Two points clarify the ``step0'' label. First, ``step0'' names
how the site is \emph{identified}---by the dominant routing weight at the direct single-hop step---not a
step-specific weight: because the backbone and memory are weight-tied across iterations, editing this expert's
value row edits the same parameters that every iteration rereads. Second, this is why suppressing the expert at
step0 only partially degrades recall ($0.380$--$0.939$) whereas suppressing it at every iteration collapses it
($0.020$--$0.460$): the fact is reread across steps, so the edited weight matters at each reread, not at step0
alone.

\paragraph{Same-location reuse.}
On every tested atom of each of the five seeds (Appendix~\ref{app:localization-protocol}),
the expert used by the direct single-hop fact at step0 is the same expert used when
that fact is reread as the second-hop fact at step1 in a 2-hop query. Thus composition rereads the same stored location
rather than consulting a separate 2-hop copy.

\paragraph{Iterative single-hop connectivity.}
A coarse symmetric bridge-interchange diagnostic at four candidate sites (one per loop iteration, $R{=}4$;
replacing the residual at the answer-position
after each loop iteration with the donor's residual) yields a best-site flip-to-donor rate of $1.00$ on
each of the five LMC seeds (best site $=$ step$1$ throughout; per-seed numbers in
Table~\ref{tab:interchange-perseed}). The finer per-pair bridge-substitution measurement, which scores
flips at the bridge-entity token specifically rather than at the loop-iteration boundary, gives $92$--$99.5\%$
flips across five seeds with any-change rate $100\%$ (i.e.\ every intervention changes the
answer away from the original, even when it does not land exactly on the donor answer). The mechanism is the iterative single-hop
composition described in \S\ref{sec:task}: step0 resolves the first-hop fact, step1 rereads the bridge entity
and resolves the second-hop fact. This is
precisely the setting in which editing the single stored first-hop fact should propagate if the memory site is
indeed reused.

\section{Study 2: Addressable Memory Affords Edit Propagation}
\label{sec:study2}

Study~2 asks whether a successful direct edit propagates through the composition that normally rereads the
edited fact. We first give the cross-substrate result, because it is the paper's main adjudication, and then show
that the LMC memory edit used in that comparison is precise and local.

\paragraph{Two measurement sets.}
We use two distinct probe sets and never mix their numbers. The \emph{cross-substrate shared-ID set} is the
intersection of facts that the three edit-comparison substrates (\runA{}, \runP{}, \runD{}) answer correctly
in-distribution before any edit; it is the only fair basis for the cross-substrate adjudication, because it holds
the evaluated facts fixed across substrates that otherwise memorize slightly different subsets. \runM{}, added
as the no-loop control, is scored on this same fixed set; as for every cell its propagation rate is conditioned
on its \emph{own} direct edit success (\S\ref{sec:editprotocol}), so any shared-set fact that \runM{} does not
answer or edit successfully pre-edit is excluded from its rate rather than counted as a propagation failure. All
Table~\ref{tab:propagation-ladder} and Table~\ref{tab:app-sot-prop} numbers are on this set. The
\emph{LMC-internal probe set} is LMC's own high-confidence pre-edit-correct set (the gold-probability
${\geq}0.90$, top-$2$ margin ${\geq}0.50$ gate of Appendix~\ref{app:localization-protocol}); it is used only to
characterize how precise an LMC edit is in isolation (Table~\ref{tab:lmc-edit}) and never enters a
cross-substrate comparison. The two sets give different LMC numbers, for a reason we make explicit: the shared
set applies only the correct-on-all-substrates criterion, \emph{without} the ${\geq}0.90$ confidence gate of the
LMC-internal set, so it admits facts at which LMC is correct but lower-confidence---exactly the facts at which a
localized edit occasionally fails to take or whose second-hop reread is less reliable. This is why LMC direct
edit success is $100\%$ on the internal set but $177$--$229$ of $213$--$247$ probes on the shared set, and why
intended propagation is $0.989$ on the internal set but $0.78$--$0.92$ on the shared set. The shared-set numbers
are the conservative ones and carry the paper's claims; the internal-set numbers are a within-LMC precision
ceiling.

\subsection{Cross-substrate propagation ladder}
\label{sec:three-substrate-results}

All four cells are edited under the common protocol from \S\ref{sec:editprotocol}: rates are measured on a
shared pre-edit-correct ID set and conditioned on direct atomic edit success.

\paragraph{Edit-site fairness across substrates.} The four configurations are not edited at the same physical
weights, because they do not have the same fact-bearing site: the two memory-bearing cells (LMC and Dense+Mem)
are edited at a single value row of their dominant memory expert (the site identified in \S\ref{sec:bridge-results}),
whereas the memory-free Loop and Dense are edited at value columns of the corresponding MLP output projection
(\texttt{c\_proj})---the standard targeted-edit site for dense
transformer MLPs in the ROME/MEMIT line. We use each architecture's most natural edit site rather than
imposing a single physical site, because the comparison of interest is whether \emph{after} a successful
localized edit the substrate's downstream computation can reread it, not whether one shared physical recipe
fits all backbones. Two facts limit how much this could distort the comparison. First, Loop and Dense reach
$100\%$ direct edit success on every seed, and LMC's conditioned direct success stays in
$177$--$229$ of $213$--$247$ probes (lowest rate is s2 at $184/247{=}0.745$ on the
shared cross-substrate set, see Table~\ref{tab:app-sot-prop}), so the chosen sites are causally effective at flipping the single-hop
fact. Second, the propagation gap separates configurations along the memory axis (LMC and Dense+Mem both
propagate strongly; Loop and Dense do not), \emph{even though the two memory-bearing cells are edited at an
expert value row while the two memory-free cells are edited at \texttt{c\_proj} columns}---so the gap is not
explained by the choice of edit operator (expert-row vs.\ \texttt{c\_proj}) alone, since Dense+Mem and Loop
fall on opposite sides of the split despite Dense+Mem sharing the dense backbone family. We return to the residual concern (whether a coarser-grained memory would
still afford propagation) in \S\ref{sec:discussion}.

\begin{figure}[t]
\centering
\includegraphics[width=0.78\linewidth]{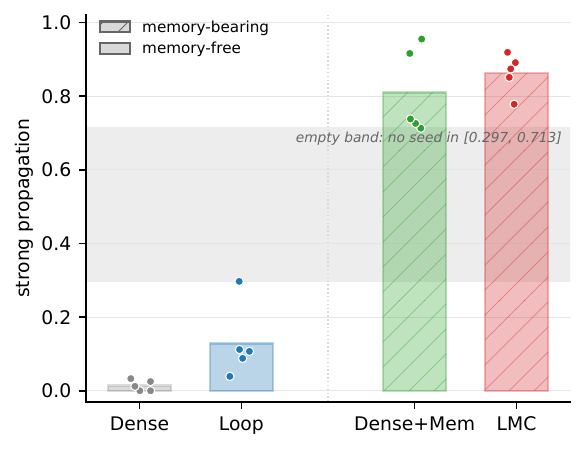}
\caption{\textbf{Edit-propagation ladder by substrate.} Bars show five-seed means of the strong-propagation metric; dots show individual seeds. Colors follow the Study~1 substrate palette, and hatching marks memory-bearing cells. The grey band is empty: every memory-bearing seed is at least $0.713$, while every memory-free seed is at most $0.297$. Thus the main split is along the memory axis, not the recurrence axis; Loop is a weak intermediate rather than a second high-propagation cell.}
\label{fig:propagation-ladder}
\end{figure}

Crucially, the low-propagation
substrates are not an artifact of failed edits: Loop and Dense reach \emph{100\% direct edit success on every
seed}, and conditioned LMC direct success stays high ($177$--$229$ of $213$--$247$ probes; per-seed values in Table~\ref{tab:app-sot-prop}). The
separation therefore reflects whether a successful direct edit propagates, not whether the edit takes.
Read along the memory axis, the result is clear: \runM{} propagates as strongly as \runD{} despite having
no loop recurrence at all, while the looped-but-memory-free \runP{} propagates only weakly. Recurrence is
therefore neither necessary nor sufficient for edit propagation; an addressable memory that the forward
computation rereads is what matters. (We do not read the \runD{}${\approx}$\runM{} non-significance as
equivalence; at $n{=}5$ vs.\ $n{=}5$ a moderate gap is not excluded, see \S\ref{sec:discussion}.)
Figure~\ref{fig:propagation-ladder} and Table~\ref{tab:propagation-ladder} show the strong-propagation ranges. The two memory-bearing
configurations propagate strongly (LMC $0.778$--$0.919$, mean $0.863 \pm 0.053$; Dense+Mem $0.713$--$0.955$, mean
$0.810 \pm 0.116$), Loop remains intermediate ($0.039$--$0.297$, mean $0.129 \pm 0.098$), and Dense stays
near zero ($0.000$--$0.033$, mean $0.014 \pm 0.015$). We report the separation descriptively because with
$n{=}5$ vs.\ $n{=}5$ the exact two-sided Mann--Whitney test attains its floor under complete rank separation:
every memory-bearing seed exceeds every memory-free seed, no memory-free seed reaches even the lowest
memory-bearing seed (empty band $[0.297,0.713]$), and the ladder does not rest on any single outlier. For
reference, this complete separation gives $U{=}25/25$ and the minimum attainable two-sided $p{=}0.0079$.
We do not detect a difference between the two memory-bearing cells at this sample size (LMC vs
Dense+Mem, $U{=}16/25$, $p{=}0.55$; bootstrap difference of means $0.05$, $95\%$ CI $[-0.05, 0.15]$, straddling
zero); we do not read this as equivalence. The data thus support a two-level addressability split, not
a four-way ranking.\footnote{Tests are over the five per-seed strong-propagation values per cell
(Table~\ref{tab:app-sot-prop}). We use the exact (enumerated) two-sided Mann--Whitney $U$ test; for $n{=}5$ vs
$n{=}5$, complete rank separation gives $U{=}25/25$ at the minimum attainable two-sided $p{=}0.0079$. Bootstrap
CIs resample seeds with replacement ($2\times10^{5}$ draws).}

\paragraph{The split is quantitative, not definitional.}
It might seem tautological that a memory with separately-writable sites supports localized edits while a substrate without them does not. Three things make the result non-trivial here. (i) The split is quantitative: memory-bearing cells propagate at $0.71$--$0.96$ versus \runP{}'s $0.04$--$0.30$ and \runA{}'s near-zero $0.00$--$0.03$, with complete rank separation across seeds, and \runP{} is weak but nonzero, so the operative variable is not the bare presence of a slot but an addressable site that the forward computation writes to and later rereads. We flag that \runP{}'s status as a true \emph{intermediate} (rather than effectively memory-free) rests largely on one seed (s2 at $0.297$; the other four are $0.039$--$0.112$, near \runA{}); at $n{=}5$ we therefore treat ``Loop is intermediate'' as suggestive, and rest the main claim on the memory/no-memory split, which holds for every seed. (ii) The advantage persists when the loop is removed entirely (\runM{}), so it is not a property of recurrence. (iii) It also survives coarsening the store to $N{=}13$ (granularity sweep, \S\ref{sec:discussion}), so it is not an artifact of our chosen granularity. We return to this point in \S\ref{sec:discussion}.

\begin{table}[t]
\centering
\caption{Main edit-propagation result (five-seed strong-propagation ranges, conditioned on direct edit
success). The grid crosses backbone (dense vs.\ looped) with shared memory (absent vs.\ present). The split is
along the memory axis: both memory-bearing configurations propagate strongly and show no detectable difference
(LMC vs.\ Dense+Mem, $U{=}16/25$, $p{=}0.55$), both cleanly separate from the two memory-free
configurations (complete rank separation; $U{=}25/25$, minimum two-sided $p{=}0.0079$), and the looped-but-memory-free configuration is only an intermediate,
partial case. Full per-seed values are in Appendix~\ref{app:full-precision}.}
\label{tab:propagation-ladder}
\small
\begin{tabular}{lcc}
\toprule
Backbone & no shared memory & shared memory \\
\midrule
dense  & Dense: $0.000$--$0.033$ & Dense+Mem: $0.713$--$0.955$ \\
looped & Loop: $0.039$--$0.297$  & LMC: $0.778$--$0.919$ \\
\bottomrule
\end{tabular}
\end{table}

This ladder rules out a flat shared-access account in which LMC and Loop both propagate strongly, and refines
the simpler addressability-only claim that only memory matters: Loop is a weak, partial case rather than fully
memory-free. We do not claim the \emph{direction} of this result is surprising---that an edit to an explicit,
addressable store reread downstream propagates, while a distributed dense-weight edit does not, is the working
intuition behind the ROME/MEMIT/MQuAKE line. What this study adds is a controlled, quantified
dissociation: a $2{\times}2$ grid that holds effective depth fixed, a no-loop memory control (Dense+Mem) showing
recurrence is neither necessary nor sufficient for propagation, and a looped no-memory control (Loop) that
places recurrence as at most a partial route. Dense weak
changes mostly reflect nonspecific disruption of memorized ID answers rather than propagation to the edited
bridge (Appendix~\ref{app:weak-prop}); on a matched held-out unrelated-fact set the substrate differences are
specifically in propagation, not in collateral damage---all three substrates move ${\lesssim}1.5\%$ of unrelated
facts (Appendix~\ref{app:weak-prop}, Table~\ref{tab:specificity}). Full per-seed values and the 2$\times$2 table are in
Appendix~\ref{app:full-precision}.

\subsection{LMC one-row edits are precise}
\label{sec:lmc-edit-results}

The LMC result is not obtained by a broad destructive write. At the step0 edit site for the
target atomic fact, we edit only the single largest activated value row ($m{=}1$). This edit
changes the direct fact with 100\% success and propagates to bridge 2-hop queries with mean propagation
$0.989$.

\begin{table}[t]
\centering
\caption{LMC step0 $m{=}1$ value-row edit at the dominant memory site, evaluated on the
\emph{LMC-internal} probe set (atoms LMC predicts pre-edit, with their bridge 2-hop queries). These are
within-LMC precision measurements; the cross-substrate comparison uses the shared-ID set (see ``Two measurement
sets'' above; Table~\ref{tab:propagation-ladder}, Table~\ref{tab:app-sot-prop}).}
\label{tab:lmc-edit}
\small
\begin{tabular}{lcccc}
\toprule
Seed & Direct success$^{\dagger}$ & Bridge 2-hop prop.$^{\dagger}$ & Random 1-hop moved & Random 2-hop moved \\
\midrule
s0 & 1.000 & 0.995 & 0.0013 & 0.0000 \\
s1 & 1.000 & 0.988 & 0.0013 & 0.0013 \\
s2 & 1.000 & 0.988 & 0.0014 & 0.0014 \\
s3 & 1.000 & 0.993 & 0.0013 & 0.0033 \\
s4 & 1.000 & 0.979 & 0.0000 & 0.0020 \\
\bottomrule
\end{tabular}
\par\smallskip
\footnotesize $^{\dagger}$ LMC-internal probe set; the cross-substrate comparison uses the shared-ID set (Table~\ref{tab:app-sot-prop}).
\end{table}

Additional edit-budget sweeps clarify what the $m{=}1$ edit actually buys. Direct success ($1.000$) and intended
bridge propagation ($\approx0.99$) are already saturated at $m{=}1$ and do not improve as $m$ grows to the full
$320$ rows; what changes is collateral movement of \emph{co-resident} facts---other facts routed to the
same dominant expert as the edited fact---which rises roughly an order of magnitude (from $0.06$--$0.30$ at
$m{=}1$ to $0.78$--$0.92$ at $m{=}320$; Appendix~\ref{app:msweep}). (This collateral range $0.78$--$0.92$ coincides numerically with the \emph{intended} cross-substrate propagation range of Table~\ref{tab:propagation-ladder} by accident; here it is unwanted movement of co-resident facts, there it is wanted propagation to the bridge query.) The $m{=}1$ edit is therefore not required
for the fact to \emph{write or propagate}; its value is a nearly free order-of-magnitude reduction in collateral,
i.e.\ locality bought at essentially no cost in direct success or intended propagation. The locality result has
one caveat: these co-resident facts, sharing the same expert, can move. Leakage increases with hidden-state cosine, from about 5\%
in the $[0.0,0.2)$ bin to 21\% in $[0.2,0.4)$ and 50\% in $[0.4,0.6)$, while random unrelated facts are nearly
unchanged.

\subsection{Auxiliary timing diagnostic}
\label{sec:asp-results}

We also ran an answer-subspace-preserving interchange (HOLDANS; Appendix~\ref{app:asp-calibration}) to probe
when usable bridge information first appears. It orders the three substrates in the same direction as
propagation (LMC$\,>\,$Loop$\,>\,$Dense) but is not a seed-level predictor: it saturates at $1.0$ on every LMC
seed and, within Loop, is decoupled from---even anti-correlated with---propagation (the highest-HOLDANS Loop
seed propagates least). We therefore treat it as a coarse, substrate-level observation consistent with the
ordering, not as a mechanism, and do not rely on it for any claim. The intervention and full per-seed values
are in Appendix~\ref{app:asp-calibration} (Table~\ref{tab:asp-interchange}).

\section{Related Work and Positioning}
\label{sec:related}

\paragraph{Looped and recurrent-depth transformers.}
A growing family of models obtains additional computation by iterating a shared set of layers in
the depth direction (``vertical recurrence'') rather than by adding parameters. The Universal Transformer
\citep{dehghani2019universal} first proposed weight-tied
recurrence; Huginn
\citep{huginn} scales recurrent depth to $3.5$B parameters with a $\textit{prelude}$--$\textit{recurrent}$--$\textit{coda}$
structure and randomly sampled iteration counts, while Ouro \citep{ouro} loops its \emph{entire}
layer stack ($24$ layers for the $1.4$B model, $48$ for $2.6$B) $R{=}4$ times and argues that
recurrence improves knowledge \emph{manipulation} rather than capacity. On the theoretical side,
looped transformers have been shown to length-generalize on algorithmic tasks by emulating
iterative computation \citep{fan2024looped}, and \citet{saunshi2025looped} show that looped models
carry an inductive bias toward \emph{reasoning}---matching far deeper iso-FLOP baselines on
reasoning primitives despite worse perplexity than comparable-size dense models, a gap rooted in
the long-known capacity cost of cross-layer parameter sharing
\citep{lan2020albert,takase2021sharing,csordas2024moeut,mohtashami2025cotformer}. LMC sits at the
extreme shallow end of this spectrum: only $3$ unique layers, looped $R{=}4$, with \emph{no}
prelude/coda. Unlike these models, LMC adds an explicit external memory queried between iterations,
and we use the looped-vs-dense contrast as a controlled variable rather than as the headline result.

\paragraph{Grokking of implicit reasoning.}
\citet{power2022grokking} first reported grokking---delayed generalization long after the training
loss saturates---on small algorithmic tasks, where generalization is essentially binary: validation
accuracy eventually climbs from chance to near-perfect. \citet{nanda2023progress} and
\citet{varma2023circuit} analyze the transition mechanistically, explaining it as a slow, efficient
generalizing circuit overtaking a fast memorizing one. \citet{wang2024grokked} carry grokking into
knowledge-graph reasoning: a vanilla transformer learns the in-distribution task but never
generalizes OOD to held-out two-hop compositions, and their analysis predicts that cross-layer
memory-sharing---via either memory augmentation or explicit recurrence---is needed for OOD
systematicity. That prediction is not itself a controlled separation; concurrent work
\citep{kohli2026loopthink} confirms the loop-only half on the same task. \textbf{Our contribution is
orthogonal: building on Wang's task design and recipe, we separate two implementations of repeated shared
access---loop recomputation and memory rereading---and then ask whether they remain equivalent under
edit-propagation interventions.}

\paragraph{Memory-augmented models.}
Separating a controller from an external memory is a long-standing goal: Neural Turing Machines and
Differentiable Neural Computers couple controllers to addressable memories
\citep{graves2014ntm,graves2016dnc}, and Memory Networks retrieve supporting facts for question answering
\citep{weston2015memory,sukhbaatar2015endtoend}. Modern transformer memories take several forms---segment-level
memory tokens \citep{bulatov2022rmt}, non-parametric caches \citep{wu2022memorizing}, trainable key--value
stores \citep{berges2024memorylayers,tu2025mlpmemory}, and auxiliary neural memories
\citep{titans,kang2025lm2}---but most attach storage to a fixed, non-recurrent backbone. LMC differs by
\emph{coupling}: it co-trains a shallow loop and its memory from scratch, interleaving a memory retrieval into
every iteration rather than doing long-context retrieval. This coupling matters: \citet{geng2024knnlimits} show
that bolt-on $k$NN-LM memory helps memory-intensive tasks yet \emph{hurts} multi-hop reasoning even with perfect
retrieval, so that our in-loop memory instead \emph{accelerating} the reasoning grokking transition is a
non-trivial consequence of how it is coupled, not of adding a store per se.

\paragraph{Model editing and edit propagation.}
ROME and related causal-tracing work localize factual associations in dense transformers and edit them with
rank-one updates \citep{meng2022rome}; MEND and MEMIT scale this direction to faster or batched factual
edits \citep{mitchell2022mend,meng2023memit}. Multi-hop editing benchmarks such as MQuAKE show that a
successful direct edit often fails to propagate through downstream compositions \citep{zhong2023mquake}.
Our goal is a controlled model organism for edit propagation: the fact-bearing site is causally localized, the
normal computation either rereads it or does not, and propagation is measured after conditioning on direct edit
success.

\paragraph{Concurrent loop-plus-memory studies.}
Closest to us are recent works pairing recurrence with memory. \citet{kohli2026loopthink}, on the
\emph{same} knowledge-graph task as ours, show that a looped transformer groks where a vanilla one
fails---but study only the loop route (varying the recurrence count $R$), with no external memory and
no loop-vs-memory separation; their looped comparison also lets effective depth grow with $R$
($4\!\to\!32$), entangling the recurrence \emph{mechanism} with added depth, whereas we hold effective
depth fixed. \citet{frey2026loopsmem} augment an adaptively looped language model with gated
local/global memory banks, and \citet{sapunov2026utmem} add learned memory tokens to a single-block
Universal Transformer on Sudoku; both separate knowledge \emph{manipulation} (looping) from knowledge
\emph{capacity} (memory), whereas we separate the \emph{two routes to cross-layer sharing} themselves
(loop-recompute vs.\ memory-reread) under a shared task and intervention protocol. Our memory is also a
sparse mixture-of-experts \emph{knowledge store} consulted between iterations---holding retrievable
\emph{knowledge}, not the transient computation state of a scratchpad/register
\citep{sapunov2026utmem} or per-layer gated key--value bank
\citep{frey2026loopsmem}.\footnote{\citet{sapunov2026utmem} find memory and ponder depth
\emph{substitutable}, while \citet{frey2026loopsmem} find them \emph{complementary}; our controls
show that either route alone can trigger the transition, but the degree of substitutability depends on
the scale and allocation being compared.}

\section{Discussion and Limitations}
\label{sec:discussion}

\paragraph{Computation and storage are separable.}
The claim is about decoupling rather than recurrence per se: a fact-bearing store can be paired with
different computation modules. Both Dense+Mem (a single-pass dense backbone) and LMC (a shallow looped
backbone) instantiate this decoupling and both grok, so the transition does not require a recurrent module.
The loop is simply a frugal choice---fewer unique parameters than the dense backbone---and, at matched scale
(\S\ref{sec:axis2runs}), it is the faster of the two to reach OOD onset.

\paragraph{The dense-backbone memory cell isolates the site from the loop.}
As shown in \S\ref{sec:three-substrate-results}, the key control is Dense+Mem: no loop recurrence, yet it
propagates edits as strongly as LMC ($0.713$--$0.955$ vs.\ $0.778$--$0.919$) with no detectable difference
from it at our sample size (Mann--Whitney $U{=}16/25$, $p{=}0.55$). We do not read this
non-significance as equivalence: with $n{=}5$ vs.\ $n{=}5$ the test is underpowered, and the bootstrap
difference of $0.05$ has a $95\%$ CI of $[-0.05, 0.15]$, so a moderate gap of order $0.15$ cannot be ruled
out. Both memory cells, however, show complete rank separation from the two non-memory cells. The propagating
factor is an editable site that an edit can write to and that later computation rereads, not loop recomputation
per se. Loop is an intermediate, partial case rather than a co-equal route: it shares computation but exposes no
separately-writable site, so localized edits sometimes propagate but mostly do not. A timing diagnostic
(Appendix~\ref{app:asp-calibration}) is consistent with this substrate-level ordering but fails at the seed
level, so we do not use it to explain individual seeds; the operative distinction remains whether the substrate
exposes an addressable site, not when content becomes usable.

\paragraph{Why the addressability split is not definitional.}
As argued above, the split is quantitative rather than definitional: Loop is weak but nonzero (a partial intermediate, on the caveat of \S\ref{sec:three-substrate-results}), the memory
advantage persists when recurrence is removed entirely (Dense+Mem), and the granularity sweep below shows that
it is not an artifact of using the fine-grained $N{=}128$ store.

\paragraph{Measurement and scope.}
Because Dense does not answer OOD compositions correctly, OOD propagation would conflate edit propagation with
basic compositional failure; we therefore use a shared pre-edit-correct ID set for the edit-propagation
comparison while retaining OOD as the Study~1 learning metric. More broadly, our experiments are deliberately
controlled and do not by themselves establish that the same affordance appears in pretrained large language
models or retrieval-augmented systems. They establish a mechanism precondition: if a model stores facts in an
addressable in-loop memory that the normal forward computation rereads, a localized edit can propagate through
composition in a way that dense and looped weight edits do not match. One asymmetry in our evidence should be
stated plainly: we localize facts in the LMC memory in depth (dominant expert, same-location reuse, causal
patching, and an $m$-budget sweep), but we do not run a parallel localization for Loop. Loop's weak propagation
is therefore established \emph{behaviorally}---under a top-$k$ \texttt{c\_proj} edit that reaches $100\%$ direct
success, the edit does not carry through composition---rather than by demonstrating that no separately-writable
Loop site exists. A Loop-side edit-site and budget sweep that still failed to propagate would strengthen the
``no addressable site'' reading; we leave it to future work and do not claim Loop facts are provably
non-localizable.

\paragraph{Locality and future tests.}
The LMC value-row edit is highly local for random unrelated facts, while co-resident facts can move in
proportion to hidden-state similarity---a superposition view of locality in which unrelated facts stay stable,
intended bridge updates propagate, and nearby co-resident facts reveal the remaining interference structure.
\paragraph{Granularity confound, and what an $N$ sweep tells us.}
Our editing claim is demonstrated at one memory granularity ($N{=}128$ experts, ${\approx}312$ atoms/expert),
which we deliberately chose because fine-grained routing sharpens atom-to-site addressability. This raises a
fair confound: part of LMC's editing advantage may be introduced by this design choice rather than by ``memory''
in general, so that ``memory affords editing'' risks collapsing into ``we tuned memory to be finely
addressable.'' Because the main comparison fixes $N{=}128$, we run a granularity sweep below to
separate the role of addressing granularity from the presence of memory. The mechanism analysis
already suggests \emph{why} granularity should matter: at coarse routing ($N{=}13$, ${\approx}3077$
atoms/expert) a single atomic fact cannot occupy its own expert slot and spills onto roughly two neighbors,
whereas at $N{=}128$ a fact collapses to essentially one expert (effective fan-out ${\approx}1.04$, $95\%$
single-expert; Appendix~\ref{app:coresident} and the single-store analysis). A localized edit is only as clean
as the site it can target, so we expect the editing advantage to weaken as $N$ shrinks and the target site
smears over co-resident facts---the same superposition mechanism that already governs our co-resident leakage
result.

To check this empirically we re-ran the full three-substrate edit-propagation protocol with the LMC
configuration replaced by its coarse-granularity counterpart \runDsmall{} ($N{=}13$, same
backbone, recipe, and five seeds; see Table~\ref{tab:app-config-provenance}),
holding Loop and Dense fixed; all numbers below are on the same cross-substrate shared-ID set as
\S\ref{sec:three-substrate-results} (Table~\ref{tab:app-sot-prop}), so the $N{=}128$ and $N{=}13$ figures are
like-for-like. The sweep splits cleanly along the paper's own editability-vs-propagation axis. \emph{Editability
is preserved}: direct edit success is essentially unchanged ($N{=}128$ mean ${\approx}0.90$, range
$0.75$--$1.00$; $N{=}13$ $0.84{\pm}0.15$, fully overlapping), so coarsening the store does not stop a localized
edit from taking. \emph{Propagation attenuates but does not collapse}: LMC strong propagation moves from
$0.863$ (range $0.778$--$0.919$) at $N{=}128$ to $0.78{\pm}0.09$ at $N{=}13$---its $N{=}13$ mean sits at the
$N{=}128$ worst-seed value ($0.778$)---while the memory-free controls do not move (Loop $0.13{\pm}0.10$, Dense
$0.01{\pm}0.02$, both indistinguishable from $N{=}128$). Crucially the substrate separation survives at coarse
granularity: the $N{=}13$ LMC strong-propagation distribution remains fully disjoint from Loop and Dense
(complete rank separation; exact Mann--Whitney $U{=}25/25$, minimum two-sided $p{=}0.0079$ for LMC vs.\ Loop and for LMC vs.\ Dense, matching the $N{=}128$
separation), at roughly $5{\times}$ and ${\approx}40{\times}$ their rates, so the two-level memory/no-memory
split is not an artifact of fine granularity. What granularity buys is precision, not the existence of the
affordance: at $N{=}128$ a fact collapses to essentially one expert and the edit is near-surgical, whereas at
$N{=}13$ (${\approx}3077$ atoms/expert) the target site smears over co-resident neighbors and propagation
degrades by the same superposition mechanism that governs our co-resident leakage result
(Appendix~\ref{app:coresident-leakage}). The claim is therefore refined, not retracted: an addressable, in-loop
memory affords edit propagation that dense and loop-only substrates do not, and the precision of that
propagation scales with addressing granularity. The narrow form, ``a \emph{fine-grained}, addressable memory
affords edit propagation,'' is supported with highest precision; the broader form, ``addressable memory affords
edit propagation,'' holds in direction at $N{=}13$ at a measured granularity cost.

\paragraph{Falsifiable $N$-hop predictions (future work).}
Having shown the claim survives a granularity sweep, we note that the addressable-site reading also suggests
falsifiable predictions for longer compositions that we leave to future work. If propagation depends on how much
computation remains to reread the edited site, fixed-depth Loop should decay faster with hop length than LMC,
and adding loop iterations after the editable site is read should recover some propagation. We do
not run a 3-hop point in this paper: the present scope deliberately stops at 2-hop, where the bridge site and
the reread step are both causally identified, and a clean $N$-hop replication would require redesigning the
data-generation protocol, re-establishing grokking at the new hop length, and re-localizing the bridge
site---a separate study rather than a single supplementary point.

\section{Conclusion}
\label{sec:conclusion}

Looped transformers and memory-augmented looped transformers both solve an OOD compositional grokking problem
that dense no-memory transformers fail. The common enabling ingredient is repeated shared access: recomputing
with a tied backbone and rereading a shared store are two ways to reuse a fact-bearing substrate across a
computation. But intervention reveals a difference. Fine-grained LMC memory localizes atomic facts to
addressable sites, rereads the same site during iterative single-hop composition, and supports one-row edits that propagate
through 2-hop reasoning with low unrelated collateral. In the edit grid, both memory-bearing
cells propagate strongly (Dense+Mem and LMC), Loop remains intermediate, and Dense remains near zero.
Loop and memory both cross the OOD grokking barrier that the dense no-memory model never does; at
matched scale the loop route is in fact the faster of the two to onset (\S\ref{sec:axis2runs}), so even
for learning the routes are not symmetric. For editing the asymmetry reverses. What matters is addressability: whether the substrate exposes a site that an edit can target and that
later computation rereads. The two memory-bearing cells provide such a site and propagate; the dense
no-memory model provides neither and does not; loop---which reuses a fact-bearing computation but for which we do not identify a comparable
separately-writable site under our edit protocol---is a partial, intermediate case.
The precision of propagation, in turn, scales with how finely the store is addressed.

\clearpage
\appendix

\section{Model, configuration, and training details}
\label{app:config}

This appendix starts with the information a reviewer needs to reproduce the controlled comparisons: the primary
architecture grid, the auxiliary controls, and the shared optimizer/data recipe.

\paragraph{Reproducibility.} The synthetic KG-QA dataset is generated procedurally from the
parameters in Table~\ref{tab:app-training} (2000 entities, 200 relations, per-entity out-degree 20, $2$-hop
inference ratio $\phi_r{=}12.6$, word-level vocabulary of $2202$) with a fixed generation seed, so the full
corpus is reconstructible from the released generator. The model definitions, data-generation and tokenization
scripts, training configurations for every cell and seed, and the localization/edit-propagation evaluation
harness will be released as an anonymized repository accompanying the submission and de-anonymized on
acceptance. All main claims are reported over five seeds (s0--s4), with per-seed values in the appendices.

\paragraph{Ethics and broader impact.} This is a controlled interpretability study on synthetic data with no
human subjects and no released model capable of real-world factual recall. We study edit propagation to
understand when localized edits behave predictably; we do not propose a deployment editing method.

\begin{table}[h]
\centering
\caption{Configuration dictionary organized around the 2$\times$2 control design. The primary cells cross
backbone type (dense versus looped) with shared-memory access (absent versus present). The auxiliary controls
test whether the conclusion depends on sparse routing or full weight sharing.}
\label{tab:app-config}
\small
\resizebox{\linewidth}{!}{%
\begin{tabular}{lllll}
\toprule
Name & Backbone & Memory & Access pattern & Role \\
\midrule
\runA{} & 12L dense & none & none & anchor / neither factor \\
\runP{} & 3L$\times$4 loop & none & repeated recomputation & loop-alone cell \\
\runM{} & 12L dense & shared memory & repeated memory reads & dense memory-only control \\
\runD{} & 3L$\times$4 loop & shared memory & recompute + reread & LMC / loop+memory cell \\
\midrule
Shared-FFN memory control & 3L$\times$4 backbone & shared dense FFN & 3 repeated reads & no sparse-routing control \\
Partial-sharing loop control & 4L$\times$3 loop & none & 3 recurrent applications & reduced-sharing loop control \\
\bottomrule
\end{tabular}%
}
\end{table}

\begin{table}[h]
\centering
\caption{Shared training configuration. All runs use the same optimizer, schedule, and data so that the only
varied factors are backbone type and shared-memory access. Effective depth is held at 12 across the primary
cells (dense 12L; looped 3L$\times$4). The fine-grained LMC uses $N{=}128$ routed value experts with top-$k{=}2$
and per-expert hidden $320$, matched to iso-total memory parameters.}
\label{tab:app-training}
\small
\begin{tabular}{ll}
\toprule
Setting & Value \\
\midrule
Backbone width $d_{\mathrm{model}}$ & 768 \\
Attention heads & 12 \\
MLP ratio & 4 \\
Effective depth & 12 (dense 12L; loop 3L$\times$4, $R{=}4$) \\
Position embedding & learned absolute \\
Optimizer & AdamW ($\beta_1{=}0.9$, $\beta_2{=}0.95$) \\
Learning rate & $1\times10^{-4}$ constant (2000-step warmup) \\
Weight decay & 0.1 \\
Gradient clip & 1.0 \\
Batch size & 512 (no gradient accumulation) \\
Precision & bfloat16 \\
Max steps & up to $5\times10^{6}$ (until grok or convergence) \\
Seeds & 5 (s0--s4) \\
Memory (LMC $N{=}128$) & 128 experts, top-$k{=}2$, expert hidden 320 \\
Vocabulary size & 2202 (word-level) \\
KG entities / relations & 2000 / 200 \\
Per-entity out-degree & 20 \\
Atomic facts (total) & ${\approx}40{,}000$ ($2000\times20$); recall probe uses a $4{,}000$ subset \\
2-hop inference ratio $\phi_r$ & $12.6$ (composed 2-hop / atomic), Wang App.\ E.2 setting \\
Hop structure & $2$-hop only ($n_{\min}{=}n_{\max}{=}2$) \\
Data & synthetic KG-QA \citep{wang2024grokked}: atomic $(e,r)\mapsto e'$ \\
 & and 2-hop $(e_0,r_1,r_2)\mapsto e_2$; train on single-hop + ID, \\
 & evaluate on held-out OOD compositions \\
\bottomrule
\end{tabular}
\end{table}

\begin{table}[h]
\centering
\caption{Run dictionary for the \textsc{LMC} family. \textsc{LMC} denotes the same
loop-plus-routed-memory \emph{role} throughout, but four runs instantiate it at different scales,
granularities, and measurement sets, so we give each a distinct code and use that code wherever the
run appears. This table is the single source of truth: the primary four-cell \runD{} and the
same-scale \runDiso{} are different configurations (hence $45$--$65$k vs.\ $166$k steps-to-onset),
and cross-substrate numbers are compared only to other cross-substrate numbers. ``${\approx}87$M-class'' refers to the parameter scale of the primary \runD{} cell; the four cells of the main grid are \emph{not} parameter-matched (that is the job of the \runDiso{}/\runPbig{}/\runMiso{} same-scale controls), so the label marks order-of-magnitude, not an exact match. \texttt{D11} is the internal config code for the $N{=}13$ coarse-granularity run.}
\label{tab:app-config-provenance}
\small
\resizebox{\linewidth}{!}{%
\begin{tabular}{lllll}
\toprule
Code & Where used & Configuration & Scale & Measurement set \\
\midrule
\runD{} & \S\ref{sec:study1}--\S\ref{sec:study2} (primary) & 3L$\times$4 loop $+$ $N{=}128$ top-$k{=}2$ memory & ${\approx}87$M-class & four-cell grid; cross-substrate shared-ID; LMC-internal \\
\runDiso{} & \S\ref{sec:axis2runs} (Axis-2 same-scale) & same role, scale-matched to \runPbig{}/\runMiso{} & ${\approx}87$M (scale-matched) & grok-onset only \\
\runDsmall{} & \S\ref{sec:discussion} ($N$ sweep) & 3L$\times$4 loop $+$ $N{=}13$ memory (\texttt{D11}), same recipe & ${\approx}87$M-class & cross-substrate shared-ID \\
\runDowt{} & Appendix~\ref{app:owt} & LMC widemem language-modeling role ($H{=}3200$, $R4$) & ${\approx}124$M & OpenWebText perplexity \\
\bottomrule
\end{tabular}%
}
\end{table}

\section{Full Study~1 per-seed grokking results}
\label{app:study1-perseed}

Table~\ref{tab:app-perseed-grok} gives the complete five-seed results behind the Study~1 summary in the main
text. The entries are intentionally reported as first-crossing steps rather than only averages, so the reader can
check both the binary grokking outcome and the across-seed spread for every shared-access control.

\begin{table}[ht]
\centering
\caption{Study~1 per-seed grokking detail. Entries are the first training step (in thousands) at which OOD
accuracy crosses $0.1$; ``never'' means OOD accuracy stayed below the threshold through the longest run. All
configurations use 12 effective layers. Every shared-access configuration groks on all five seeds with low
across-seed variance; the dense no-memory anchor never groks.}
\label{tab:app-perseed-grok}
\small
\begin{tabular}{lccccc}
\toprule
Config. & s0 & s1 & s2 & s3 & s4 \\
\midrule
Dense (no shared access) & never & never & never & never & never \\
Shared-FFN memory control & 90k & 90k & 85k & 80k & 105k \\
Partial-sharing loop control & 225k & 230k & 225k & 220k & 220k \\
Loop 3L$\times$4 & 235k & 245k & 245k & 240k & 230k \\
Dense+Mem (12L dense) & 60k & 85k & 60k & 60k & 60k \\
LMC $N=128$ & 45k & 55k & 55k & 55k & 65k \\
\bottomrule
\end{tabular}
\end{table}

\section{Site localization and edit protocol}
\label{app:edit-protocol}

This appendix specifies how facts are localized before editing, then gives the exact sparse edit update used in
Study~2.

\subsection{Localization measurements}
\label{app:localization-protocol}

\paragraph{Probe selection.} All localization and edit measurements run on an auto-selected probe set
rather than the full ${\approx}40{,}000$-fact KG, because the diagnostics are only meaningful on facts the
model already knows pre-edit. For each seed we keep an atomic single-hop fact only if (i) the model predicts
it correctly pre-edit under greedy decoding, (ii) its gold-token probability is ${\geq}0.90$ with a top-$2$
margin ${\geq}0.50$ (a high-confidence gate, so a failed edit cannot be blamed on a borderline base
prediction), and (iii) it participates in at least one held-out 2-hop composition as the first-hop fact.
Qualifying atoms are ranked by number of 2-hop probes then by base confidence and capped per seed (the
patching matrix uses ${\approx}100$ atoms/seed, the bridge-interchange diagnostic $120$ donor--recipient
pairs/seed, and the finer per-pair bridge-substitution $200$--$239$ pairs/seed; exact per-seed counts are in
the corresponding tables of Appendix~\ref{app:localization-perseed}). The selection is identical across
seeds; only the realized counts differ because each seed memorizes a slightly different high-confidence
subset.

For each atomic fact, we identify the dominant memory expert at the direct single-hop read step. The same-location
reuse test compares this expert with the expert used when the same fact appears as the second-hop fact at the
2-hop step; the match is $100\%$ on every tested atom on each of the five seeds.
The interchange-connectivity test replaces the answer-position memory input at the 2-hop step with the
corresponding memory input from another query that shares $r_2$ but has a different bridge entity.

\subsection{Edit formula}
\label{app:edit-formula}
For the first-hop fact $(e,r_1)\mapsto e_1$, the edit target is a replacement bridge entity $e_1'$ such that
$(e_1',r_2)$ is a known fact for each measured 2-hop probe. The write direction is
\begin{equation}
  d = \frac{u(e_1') - u(e_1)}{\|u(e_1') - u(e_1)\|},
\end{equation}
where $u(\cdot)$ is the output-embedding row for the answer token.

\paragraph{LMC memory write.}
For LMC, we identify the step0 dominant memory expert $a$ for the direct single-hop query and capture the
selected-expert hidden key vector $h\in\mathbb{R}^{H}$ at the answer position. The expert's output is the
key-weighted sum of its value rows, $\mathrm{out}=\sum_{j} h_j\, W_2^{(a)}[j,:]$, so we apply a
\emph{sparse masked rank-one} update restricted to the $m$ largest-magnitude rows. Let
$S=\operatorname{top}\text{-}m_j |h_j|$ be the edited row set. The update is
\begin{equation}
  W_2^{(a)}[j,:] \leftarrow W_2^{(a)}[j,:] + \alpha\,\frac{h_j}{\|h_S\|}\, d^\top
  \quad (j\in S),\qquad \text{unchanged for } j\notin S,
\end{equation}
where $h_S=(h_j)_{j\in S}$ and $\alpha$ is chosen by the direct-success sweep. This realizes
$\Delta\mathrm{out}=\alpha\, d$ exactly when the edit is unmasked ($S$ spans all rows); the paper uses the
sparsest setting $m{=}1$, editing only the single largest activated value row (so the update reduces to
$\alpha\,\mathrm{sign}(h_{j^\star})\,d^\top$ at the peak row $j^\star$). This is an edit to the same memory
site later reread by the 2-hop computation; $m{=}H$ recovers the full rank-one ROME write.

\paragraph{Loop and Dense value-column write.}
For Loop and Dense, the comparable value sites are columns of the MLP output projection (\texttt{mlp.c\_proj}). Candidate columns are ranked
by activation magnitude at the localized edit site. For selected columns $S$, the sparse write uses the same
answer direction $d$:
\begin{equation}
  \Delta W_{\ell,:,j} = \frac{\alpha}{\sum_{i\in S} h_i^2}\, h_{\ell,j}\, d
  \qquad (j\in S).
\end{equation}
The edit-propagation run uses the first budget that reaches direct success; the default sparse write
uses top-$k{=}4$ columns for Loop/Dense and sweeps scale.

\subsection{Direct-success gate and measurement}
\label{app:direct-gate}
The protocol has three gates:
\begin{enumerate}[leftmargin=1.6em,itemsep=2pt]
  \item choose $e_1'$ only when every measured $(e_1',r_2)$ fact exists;
  \item keep only compositions the model answers correctly before the edit;
  \item score strong propagation only after the direct single-hop prediction changes to $e_1'$.
\end{enumerate}
Strong propagation then requires the edited 2-hop answer to equal $\mathrm{fact}(e_1',r_2)$, not merely to
change away from the original answer.

\section{Auxiliary interchange diagnostic (not used for any claim)}
\label{app:asp-calibration}

This appendix documents an auxiliary timing diagnostic that we report for completeness but do \emph{not} use to
support any claim in the main text. It orders the substrates in the same direction as edit propagation at the
coarse three-way level but fails as a seed-level predictor (it saturates within LMC and anti-correlates with
propagation within Loop), so it functions here as an honest negative result rather than as a mechanism.

\subsection{Definition}
\label{app:asp-def}
The diagnostic is an answer-subspace-preserving interchange. The intervention compares two
residual substitutions for a recipient 2-hop query and a donor 2-hop query that share the second relation $r_2$
but have different bridge entities and answers. We define two modes:

\begin{description}[leftmargin=1.6em,itemsep=2pt]
  \item[FULL.] Replace the recipient's answer-position residual with the \emph{entire} donor residual.
  \item[HOLDANS (HOLD-ANswer-Subspace).] Replace only the donor's \emph{non-answer} content, while holding the
  recipient's answer-subspace component fixed (defined below).
\end{description}

FULL replacement injects the whole donor residual. This can be tautological late in the computation because the
donor residual may already point directly toward the donor answer. The HOLDANS interchange
removes this confound. For recipient residual $h_r$, donor residual $h_d$, recipient answer direction $u_r$, and
donor answer direction $u_d$, let $U=\operatorname{span}\{u_r,u_d\}$ and $P_U$ be the orthogonal projection onto
$U$. The intervened vector is
\begin{equation}
  h_{\mathrm{HOLDANS}} = P_U h_r + (I-P_U) h_d .
\end{equation}
Thus the recipient answer-subspace component is held fixed while the donor's non-answer residual content is
transferred. The measured event is a flip to the donor answer under this answer-preserving interchange. Such a
flip indicates that non-answer bridge information was available early enough for downstream computation to
recompute the donor answer, rather than being copied directly through the answer direction.

\begin{table}[h]
\centering
\caption{Auxiliary interchange diagnostic (per-seed HOLDANS flip rates). The flip rate tracks the coarse
cross-substrate ordering---stable and high on LMC, late and variable on Loop, and low/unreliable on Dense---but
it is \emph{not} a seed-level predictor of propagation (see note below) and is not used for any claim.}
\label{tab:asp-interchange}
\small
\resizebox{\linewidth}{!}{%
\begin{tabular}{lccc}
\toprule
Substrate & Typical site & HOLDANS flip & Interpretation \\
\midrule
LMC & step1 & 1.000 / 1.000 / 1.000 / 1.000 / 1.000 & usable bridge information present early \\
Loop & late/variable & 0.683 / 0.517 / 0.075 / 0.000 / 0.008 & late, variable bridge signal \\
Dense & no reliable & 0.000 / 0.008 / 0.042 / 0.033 / 0.092 & small flips; no strong semantic propagation \\
\bottomrule
\end{tabular}%
}
\end{table}

\paragraph{The diagnostic is a coarse, not a seed-level, predictor.}
The HOLDANS flip rate orders the three substrates in the same direction as propagation
(LMC$\approx$1 $>$ Loop$\approx$0.26 $>$ Dense$\approx$0.035), but it does not explain within-substrate variation.
LMC saturates at $1.000$ on every seed and therefore cannot follow LMC's own $0.778$--$0.919$ propagation
spread. Within Loop the seed-level flip rate is in fact \emph{decoupled} from, even anti-correlated with,
propagation: the highest-HOLDANS seed (s0, $0.683$) propagates only $0.112$, while the near-zero-HOLDANS seed
(s2, $0.075$) propagates the most ($0.297$). We therefore report this diagnostic only as a coarse,
substrate-level observation consistent with the LMC$>$Loop$>$Dense ordering, and do not use it to explain
individual seeds or to carry any claim.

\section{Full Study~2 propagation tables}
\label{app:full-precision}

This appendix collects the full Study~2 edit-propagation evidence: the substrate-by-substrate propagation
summary, the Dense+Mem memory-only control, and the per-seed values
that the main text rounds and summarizes in Table~\ref{tab:propagation-ladder}.

\begin{table}[h]
\centering
\caption{Three-substrate edit propagation, with the auxiliary HOLDANS diagnostic for reference. Strong
propagation is measured on the shared pre-edit-correct ID set and conditioned on direct edit success and carries
the paper's claims. The mean HOLDANS flip (Appendix~\ref{app:asp-calibration}) is shown only to indicate that the
auxiliary diagnostic orders the substrates in the same direction; it is not selected by propagation and is not a
seed-level predictor.}
\label{tab:three-substrate}
\small
\resizebox{\linewidth}{!}{%
\begin{tabular}{lccc}
\toprule
Substrate & Strong propagation & Mean HOLDANS flip (aux.) & Interpretation \\
\midrule
LMC & 0.778--0.919 & ${\approx}1.00$ & addressable edit; strong semantic propagation \\
Loop & 0.039--0.297 & ${\approx}0.26$ & intermediate propagation \\
Dense & 0.000--0.033 & ${\approx}0.035$ & no semantic strong propagation \\
\bottomrule
\end{tabular}%
}
\end{table}

\begin{table}[h]
\centering
\caption{Dense+Mem edit-propagation control. The protocol is the same strong-propagation measurement as the edit
comparison, with memory value-row edits applied at the first memory access of the dense backbone.}
\label{tab:m0-control}
\small
\begin{tabular}{lccc}
\toprule
Seed & Direct success & Dense+Mem strong & Dense+Mem weak \\
\midrule
s0 & 157/232 & 0.713 & 0.833 \\
s1 & 199/213 & 0.726 & 0.829 \\
s2 & 243/247 & 0.738 & 0.840 \\
s3 & 219/219 & 0.955 & 1.000 \\
s4 & 232/232 & 0.916 & 1.000 \\
\bottomrule
\end{tabular}
\end{table}

\begin{table}[h]
\centering
\caption{Per-seed strong propagation values used by the paper (three significant figures), with per-seed direct edit-success counts
(successful direct edits / usable pre-edit-correct probes). Loop and Dense reach 100\% direct success on every
seed, so their low propagation reflects failed \emph{propagation}, not failed editing.}
\label{tab:app-sot-prop}
\small
\begin{tabular}{lcccccc}
\toprule
Seed & LMC strong & LMC direct & Loop strong & Loop direct & Dense strong & Dense direct \\
\midrule
s0 & 0.891 & 228/232 & 0.112 & 232/232 & 0.000 & 232/232 \\
s1 & 0.919 & 213/213 & 0.107 & 213/213 & 0.000 & 213/213 \\
s2 & 0.778 & 184/247 & 0.297 & 247/247 & 0.033 & 247/247 \\
s3 & 0.851 & 177/219 & 0.088 & 219/219 & 0.025 & 219/219 \\
s4 & 0.874 & 229/232 & 0.039 & 232/232 & 0.012 & 232/232 \\
\bottomrule
\end{tabular}
\end{table}

\section{Weak propagation explanation}
\label{app:weak-prop}

The main text reports strong propagation: the edited 2-hop answer must equal the answer implied by the new
bridge. Weak propagation, defined as any change away from the original answer, is secondary.

\begin{table}[h]
\centering
\caption{Weak propagation rates. Dense weak changes can be high without satisfying the strong criterion:
high Dense weak rates indicate nonspecific disruption of memorized ID answers, not semantic propagation to the
answer implied by the edited bridge.}
\label{tab:app-weak-prop}
\small
\begin{tabular}{lccc}
\toprule
Seed & LMC weak & Loop weak & Dense weak \\
\midrule
s0 & 0.949 & 0.135 & 0.004 \\
s1 & 0.979 & 0.176 & 0.053 \\
s2 & 0.823 & 0.382 & 0.326 \\
s3 & 0.919 & 0.132 & 0.269 \\
s4 & 0.953 & 0.104 & 0.237 \\
\bottomrule
\end{tabular}
\end{table}

We report edit \emph{specificity} on the same footing for all substrates. Re-running the cross-substrate
edit protocol against a \emph{fixed shared held-out unrelated-fact set} ($K{=}50$ atoms, identical across
seeds and substrates), with the edit still applied we re-predict each unrelated atom's $1$-hop and $2$-hop
answer and record the fraction that \emph{moved} off its pre-edit baseline (lower is better; an ideal local
edit moves nothing). Table~\ref{tab:specificity} gives the $5$-seed result conditioned on direct success.
The picture is clean and reinforces the main reading: every substrate is highly specific to unrelated facts,
so the substrate differences in \emph{bridge propagation} are not an artifact of nonspecific damage. In
particular, LMC's strong propagation ($0.86$) is \emph{not} bought by collateral damage---all three substrates are highly specific (LMC ${\approx}1.5\%$ unrelated-fact movement, Loop and Dense ${\approx}0\%$)---and Dense's near-zero
propagation ($0.014$) coincides with near-zero unrelated movement, confirming that Dense cleanly fails to
propagate rather than thrashing nonspecifically. The weak-propagation column is therefore a within-target
measure only: the high Dense weak rate ($0.27$--$0.33$ on s2--s4) reflects nonspecific disruption of the
\emph{edited} query's memorized answer, not movement of \emph{other} facts (which is ${\approx}0$). Coarsening
the store to $N{=}13$ raises unrelated movement from ${\approx}1.5\%$ to ${\approx}5\%$---the expected
superposition cost of a coarser address (\S\ref{sec:discussion})---but it remains far below the propagation
signal, so granularity buys precision without being a precondition for the affordance.

\begin{table}[h]
\centering
\caption{Held-out unrelated-fact specificity (matched protocol, $K{=}50$ shared atoms, $5$ seeds,
mean$\pm$sd, conditioned on direct success). \emph{moved} is the fraction of unrelated atoms whose prediction
changed under the edit (lower is better). All substrates are highly specific; the substrate gap is in
\emph{bridge propagation} (right column, repeated from Table~\ref{tab:app-sot-prop}), not in collateral
damage. Coarsening LMC to $N{=}13$ (\runDsmall{}) increases leakage as predicted by superposition but stays
well below its propagation.}
\label{tab:specificity}
\small
\begin{tabular}{@{}lcccc@{}}
\toprule
substrate & unrel.\ $1$-hop moved & unrel.\ $2$-hop moved & direct success & strong propagation \\
\midrule
\runD{} ($N{=}128$)   & $0.015\pm0.002$ & $0.011\pm0.002$ & $0.905\pm0.119$ & $0.863\pm0.053$ \\
\runP{}              & $0.000\pm0.000$ & $0.000\pm0.000$ & $1.000\pm0.000$ & $0.129\pm0.098$ \\
\runA{}              & $0.000\pm0.000$ & $0.000\pm0.000$ & $1.000\pm0.000$ & $0.014\pm0.015$ \\
\midrule
\runDsmall{} ($N{=}13$) & $0.047\pm0.039$ & $0.054\pm0.035$ & $0.840\pm0.154$ & $0.779\pm0.091$ \\
\bottomrule
\end{tabular}
\end{table}

\section{Sparse edit locality and co-resident leakage}
\label{app:coresident}

\subsection{Edit-budget sweep}
\label{app:msweep}

The sparse value-row edit used in the main text writes only $m{=}1$ row at the dominant expert. To check that this $m{=}1$ choice is not driving the precision result, Table~\ref{tab:lmc-msweep} sweeps $m \in \{1,4,22,320\}$ at the same step0 site. Direct success and intended bridge propagation are already saturated at $m{=}1$ (both ${\approx}1.0$) and do not improve as $m$ grows. The single quantity that grows with $m$ is collateral movement of co-resident facts (other facts routed to the same dominant expert), which rises from $0.06$--$0.30$ at $m{=}1$ to $0.78$--$0.92$ at $m{=}320$. The $m{=}1$ choice therefore buys locality at essentially no cost in direct success or intended propagation.

\begin{table}[h]
\centering
\caption{Sparse value-row sweep at the same step0 memory site. Direct writing and intended propagation are
already saturated at $m{=}1$; increasing $m$ mainly increases co-resident collateral. Thus the $m{=}1$ edit
buys locality at almost no direct-success or intended-propagation cost.}
\label{tab:lmc-msweep}
\small
\resizebox{\linewidth}{!}{%
\begin{tabular}{lccc}
\toprule
Edited rows $m$ & Direct success & Bridge 2-hop prop. & Co-resident moved \\
\midrule
1   & 1.000 / 1.000 / 1.000 / 1.000 / 1.000 & 0.995 / 0.988 / 0.988 / 0.993 / 0.979 & 0.300 / 0.063 / 0.094 / 0.156 / 0.115 \\
4   & 1.000 / 1.000 / 1.000 / 1.000 / 1.000 & 0.996 / 0.994 / 0.994 / 0.994 / 0.988 & 0.300 / 0.375 / 0.344 / 0.188 / 0.385 \\
22  & 1.000 / 1.000 / 1.000 / 1.000 / 1.000 & 0.996 / 0.994 / 0.994 / 0.994 / 0.988 & 0.650 / 0.625 / 0.688 / 0.594 / 0.769 \\
320 & 1.000 / 1.000 / 1.000 / 1.000 / 1.000 & 0.996 / 0.994 / 0.994 / 0.994 / 0.988 & 0.900 / 0.875 / 0.812 / 0.781 / 0.923 \\
\bottomrule
\end{tabular}%
}
\end{table}

\subsection{Co-resident leakage}
\label{app:coresident-leakage}
The LMC memory edit is local with respect to random unrelated facts, but not absolutely isolated among facts
that share an expert. The pooled leakage diagnostic bins co-resident facts by hidden-state cosine with the
edited fact:
\begin{center}
\begin{tabular}{lc}
\toprule
Hidden cosine bin & moved fraction \\
\midrule
$[0.0,0.2)$ & 5\% \\
$[0.2,0.4)$ & 21\% \\
$[0.4,0.6)$ & 50\% \\
\bottomrule
\end{tabular}
\end{center}
This supports the superposition interpretation used in the discussion: unrelated facts are stable, but nearby
co-resident facts can leak in proportion to representation overlap.

\section{Language-modeling sanity check on OpenWebText}
\label{app:owt}
The KG-QA results show that decoupling can retain reasoning; this appendix asks the complementary
gatekeeping question---whether the same architectural roles remain competitive on plain language
modeling on OpenWebText (OWT). This is a separate $124$M-scale sanity check, not the $87$M-class
KG-QA control of the main text (Table~\ref{tab:app-config-provenance}). Within this OWT setting, the dense and LMC roles are matched at
comparable $\approx124$M scale, and the looped-without-memory role is the corresponding shared-
backbone ablation. The dense baseline reaches a strict validation perplexity of $21.2$; the
looped-without-memory model is higher ($26.5$), consistent with the long-known capacity cost of
cross-layer sharing \citep{lan2020albert,saunshi2025looped}; and adding the external memory
recovers most of this gap, bringing the decoupled LMC to $21.7$, within roughly half a point of
the dense baseline (Table~\ref{tab:owt-main}). The separation therefore carries a small, bounded language-
modeling cost rather than a qualitative loss (a gatekeeping prerequisite, not a perplexity win).
A deeper looped reference ($12$ layers $\times R{=}4$, not part of the main 124M-scale role
comparison) reaches $19.7$, indicating that the recurrent family's attainable perplexity is not
locked below the dense baseline when sufficient effective depth is available.
\begin{table}[ht]
  \centering
  \caption{Language modeling remains close under decoupling (OpenWebText, strict validation
  perplexity, $12$ effective layers). This table is the separate $124$M-scale OWT sanity-check
  dictionary: it reuses the KG-QA run names for architectural roles, but the parameter counts are
  not those of the main-text $87$M-class table. The dense and LMC rows are comparable at
  $\approx124$M scale; pure recurrence (\runP) costs perplexity; re-attaching the external memory
  (\runDowt) recovers almost all of it, landing within $\approx0.5$ of the dense baseline. These numbers
  serve only as a gatekeeping check that decoupling does not break language modeling; the ``loop costs
  perplexity'' gap is specific to the shallow $3$L$\times4$ configuration and closes once effective depth is
  added (Deep-Loop $12$L$\times4$ reaches $19.72$, Table~\ref{tab:owt-sweep}), so this table should not be
  read as independent evidence that memory is required.}
  \label{tab:owt-main}
  \begin{tabular}{@{}llr@{}}
    \toprule
    model & configuration & val.\ ppl $\downarrow$ \\
    \midrule
    \runA{}        & GPT-2 $12$L dense (no loop, no mem) & $21.19$ \\
    \runP{}        & looped $3$L$\times R{=}4$ (no mem)  & $26.52$ \\
    \runDowt{}     & LMC: looped $3$L$\times R{=}4$ $+$ mem & $21.73$ \\
    \bottomrule
  \end{tabular}
\end{table}

\paragraph{Memory-width and top-$k$ sweep.}
Table~\ref{tab:owt-sweep} expands the sanity check with off-grid variants. (i)~\textbf{Memory width
$H$ is not a sensitive factor in this narrow range}: comparing the matched \runDowt{} to the off-grid
$H{=}3072$ variant ($H{:}3200{\to}3072$) shifts ppl by only $0.05$. (ii)~\textbf{Recurrence depth
$R$ appears to be the larger lever}: $R{=}5$ vs.\ \runDowt{} ($R{:}4{\to}5$, $H$ fixed) gives $-0.48$
ppl. (iii)~\textbf{Sparsity is cheap in this setting}: the $H{=}3072$ top-$k{=}2$ variant vs.\ the
dense-store ($k{=}13$) variant (both $122.1$M) costs only $+0.32$ ppl, so the remaining gap to the
deeper looped reference (Deep-Loop, $19.72$) is more consistent with an effective-depth gap than
with memory sparsity alone.
\begin{table}[ht]
  \centering
  \caption{Full OWT strict validation perplexity ladder. ``total'' and ``active'' are parameter
  counts (M); the looped models use $L{=}3$ layers $\times R$; the memory models use an MoE store
  with the indicated top-$k$. \emph{Top block}: primary dense and LMC rows comparable at $12$
  effective layers and $\approx124$M scale; \runP{} is the corresponding looped no-memory ablation;
  remaining LMC rows are two single-factor ablations off \runDowt{} ($R{=}5$ and zero-init projection).
  \emph{Bottom block}: additional unmatched variants and a deeper looped reference.}
  \label{tab:owt-sweep}
  {\small
  \setlength{\tabcolsep}{4pt}
  \begin{tabular}{@{}llrrrr@{}}
    \toprule
    model & configuration & total & active & top-$k$ & ppl \\
    \midrule
    \runA{} & GPT-2 $12$L dense              & $124.4$ & $124.4$ & ---  & $21.19$ \\
    \runP{} & Looped $3$L$\times R4$ (no mem) & $60.6$  & $60.6$  & ---  & $26.52$ \\
    \runDowt{} & LMC widemem ($H{=}3200$, $R4$)  & $124.6$ & $70.5$  & $2$  & $21.73$ \\
    LMC ($R{=}5$)   & LMC widemem ($H{=}3200$, $R5$) & $124.6$ & $70.5$ & $2$ & $21.25$ \\
    LMC (zero-init) & LMC widemem ($H{=}3200$, zero-init $c_{\text{proj}}$) & $124.6$ & $70.5$ & $2$ & $21.92$ \\
    \midrule
    \multicolumn{6}{@{}l}{\emph{Additional unmatched variants:}} \\
    LMC ($H{=}3072$)      & LMC widemem ($H{=}3072$)    & $122.1$ & $70.1$  & $2$    & $21.78$ \\
    LMC (large-mem)       & LMC large-mem               & $367.6$ & $155.1$ & $4$    & $19.43$ \\
    LMC (dense store)     & LMC widemem full ($k{=}13$) & $122.1$ & $122.1$ & $13$   & $21.46$ \\
    Deep-Loop ($12$L$\times R4$) & Looped $12$L$\times R4$ (no mem) & $124.4$ & $124.4$ & --- & $19.72$ \\
    \bottomrule
  \end{tabular}
  }
\end{table}

\section{Per-seed localization measurements (LMC)}
\label{app:localization-perseed}
This appendix gives the per-seed numbers underlying the three localization measurements in
Section~\ref{sec:bridge-results}. All three measurements are reported on five LMC seeds (s0--s4)
and use the same checkpoints, measurement set sizes, and detection protocols described in the
main text.

\begin{table}[ht]
  \centering
  \caption{Per-atom routing causal-patching matrix on the LMC dominant-expert site, all five
  seeds. Each row is a single intervention condition applied to the same $\approx100$ atomic
  single-hop probes per seed. ``patch\_step0'' suppresses the detected primary expert at the
  direct-recall step only; ``patch\_all'' suppresses it at every loop iteration; ``wrong\_expert''
  and ``random\_site'' are negative controls (suppressing a non-detected expert or a random
  non-memory site respectively). Atomic recall collapses on patch\_all (mean $0.169$, range
  $0.020$--$0.460$) while staying near unity on the controls.}
  \label{tab:routing-perseed}
  {\small
  \setlength{\tabcolsep}{4pt}
  \begin{tabular}{@{}lrrrrr@{}}
    \toprule
    seed & n probes & patch\_step0 & patch\_all & wrong\_expert & random\_site \\
    \midrule
    s0 & $100$ & $0.870$ & $0.020$ & $1.000$ & $0.990$ \\
    s1 & $100$ & $0.380$ & $0.040$ & $1.000$ & $0.980$ \\
    s2 & $99$  & $0.939$ & $0.111$ & $1.000$ & $0.980$ \\
    s3 & $100$ & $0.910$ & $0.460$ & $1.000$ & $1.000$ \\
    s4 & $99$  & $0.939$ & $0.212$ & $1.000$ & $0.990$ \\
    \midrule
    mean & --- & $0.808$ & $0.169$ & $1.000$ & $0.988$ \\
    \bottomrule
  \end{tabular}
  }
\end{table}

\begin{table}[ht]
  \centering
  \caption{Coarse bridge-interchange diagnostic on LMC (one site per loop iteration, $R{=}4$), all five seeds. The diagnostic
  uses $120$ donor--recipient pairs per seed (clean correct $120/120$ on every seed). The
  per-step full-residual flip-to-donor vector is $[0.00,\,1.00,\,1.00,\,1.00]$ (ordered step0,\,step1,\,step2,\,step3) on every
  seed; the best site is step$1$ with flip-to-donor $1.00$ throughout.}
  \label{tab:interchange-perseed}
  {\small
  \setlength{\tabcolsep}{4pt}
  \begin{tabular}{@{}lrrlc@{}}
    \toprule
    seed & n pairs & clean correct & flip-to-donor by step (full) & best site / best flip \\
    \midrule
    s0 & $120$ & $120$ & $[0.00,\,1.00,\,1.00,\,1.00]$ & step$1$ / $1.00$ \\
    s1 & $120$ & $120$ & $[0.00,\,1.00,\,1.00,\,1.00]$ & step$1$ / $1.00$ \\
    s2 & $120$ & $120$ & $[0.00,\,1.00,\,1.00,\,1.00]$ & step$1$ / $1.00$ \\
    s3 & $120$ & $120$ & $[0.00,\,1.00,\,1.00,\,1.00]$ & step$1$ / $1.00$ \\
    s4 & $120$ & $120$ & $[0.00,\,1.00,\,1.00,\,1.00]$ & step$1$ / $1.00$ \\
    \bottomrule
  \end{tabular}
  }
\end{table}

\paragraph{Finer per-pair bridge-substitution.}
The five-seed per-pair bridge-substitution flip-to-donor rate spans $92$--$99.5\%$
(any-change rate $100\%$ on every seed; per-seed numbers in
Table~\ref{tab:bridgesubst-perseed}). The protocol differs from
Table~\ref{tab:interchange-perseed} in that it patches at the bridge-entity token rather than
at the loop-iteration boundary and is therefore a finer-grained test of the same
underlying mechanism.

\begin{table}[ht]
  \centering
  \caption{Per-pair bridge-substitution flip-to-donor on LMC across five seeds.
  ``flip'' is the fraction of clean-correct donor--recipient pairs whose 2-hop answer
  flips to the donor's answer when the bridge-token residual is replaced; ``any-change''
  is the fraction whose answer changes from the original. The auto-selected probe set
  differs slightly across seeds in size; flip rates lie in the $92$--$99.5\%$ band on
  every seed.}
  \label{tab:bridgesubst-perseed}
  {\small
  \setlength{\tabcolsep}{4pt}
  \begin{tabular}{@{}lrrr@{}}
    \toprule
    seed & pairs & flip-to-donor & any-change \\
    \midrule
    s0 & $200$ & $99.5\%$ & $100\%$ \\
    s1 & $199$ & $96.0\%$ & $100\%$ \\
    s2 & $239$ & $92.1\%$ & $100\%$ \\
    s3 & $239$ & $99.2\%$ & $100\%$ \\
    s4 & $238$ & $94.5\%$ & $100\%$ \\
    \bottomrule
  \end{tabular}
  }
\end{table}

\section{Auxiliary-loss diversity and the looped-refinement collapse}
\label{app:auxloss}
The step-level auxiliary-supervision note in the main text refers to a small instrumented
run in which adding a step-level shared-target next-token loss to the looped backbone collapses
the recurrence onto the final-output manifold: cosine similarity between the first-iteration and
last-iteration block-output latents rises from $0.47$ (no aux loss; the recipe used in this paper)
to $0.82$ (shared-target aux loss on every iteration). The training accuracy and OOD trajectories
are not improved by this collapse---they degrade---which is why the main configurations are
trained without step-level supervision. This is consistent with prior work on latent reasoning
\citep{simcot} once the framing is shifted: the operative variable is the \emph{diversity} of the
supervision target across iterations (here, none vs.\ a single shared next-token target), not its
presence. In particular, supervisions that vary across iterations (e.g.\ different intermediate
targets) have been shown elsewhere to help; this paper does not claim to overturn those results,
only that a homogeneous shared-target step loss undoes recurrence-as-computation.

\section{Dense random-deletion sensitivity}
\label{app:dense-sensitivity}
Raw parameter deletion is not an architecture-preserving capacity control, but it is a useful
sensitivity check. We evaluate the three available dense no-memory checkpoints (\runA{}, seeds
0--2), each with three random masks, on the same $4{,}000$ atomic single-hop facts used in
Section~\ref{sec:localization}. Table~\ref{tab:dense-sensitivity} summarizes the random deletion
curve over transformer-block parameters.

\begin{table}[ht]
  \centering
  \caption{Dense no-memory random-deletion sensitivity. Atomic recall is robust through small and
  medium random deletions, then enters a seed-dependent destructive cliff between roughly $8$M and
  $16$M deleted block parameters. This supports treating raw deletion as a sensitivity curve rather
  than a clean capacity-matched counterfactual.}
  \label{tab:dense-sensitivity}
  \begin{tabular}{@{}rrr@{}}
    \toprule
    deleted params & fraction of 64M target & atomic recall mean $\pm$ std \\
    \midrule
    $0$ & $0.000$ & $0.9985\pm0.0000$ \\
    $3.2$M & $0.050$ & $0.9949\pm0.0021$ \\
    $6.4$M & $0.100$ & $0.9607\pm0.0421$ \\
    $8.0$M & $0.125$ & $0.8942\pm0.1097$ \\
    $9.6$M & $0.150$ & $0.7773\pm0.2019$ \\
    $11.2$M & $0.175$ & $0.5933\pm0.2799$ \\
    $12.8$M & $0.200$ & $0.3721\pm0.2557$ \\
    $14.4$M & $0.225$ & $0.1941\pm0.1863$ \\
    \bottomrule
  \end{tabular}
\end{table}

The seed dependence is large in the cliff region: at $14.4$M deleted parameters, the three dense
checkpoints average $0.015$, $0.135$, and $0.432$ atomic recall, respectively. This behavior differs
from LMC memory-expert lesions at matched budgets, where $15.99$M expert deletion preserves
$0.9840$ atomic recall while reducing 2-hop OOD to $0.2606$. We therefore do not interpret raw dense
deletion as evidence for or against a clean storage module; it is an upper-bound destructive
sensitivity analysis.

\clearpage
\bibliographystyle{unsrtnat}
\bibliography{refs}

\end{document}